\documentclass[10pt,twocolumn,letterpaper]{article}

\usepackage[pagenumbers]{cvpr}
\usepackage[accsupp]{axessibility} 

\usepackage{times}
\usepackage{epsfig}
\usepackage{graphicx}
\usepackage{amsmath}
\usepackage{amssymb}
\usepackage{booktabs}
\usepackage{adjustbox}
\usepackage{makecell}
\usepackage{multirow}
\usepackage{bbding}
\usepackage{xcolor}
\usepackage{color, colortbl}
\definecolor{citecolor}{HTML}{2980b9}
\definecolor{linkcolor}{HTML}{c0392b}

\usepackage{tabularx}
\usepackage{enumitem}
\setlist[enumerate]{itemsep=0mm}
\usepackage{bbding}
\usepackage{pifont}
\usepackage{wasysym}
\usepackage{float}

\usepackage[hidelinks,pagebackref=true,breaklinks=true,colorlinks,bookmarks=false,citecolor=citecolor,letterpaper=true,linkcolor=linkcolor]{hyperref}

\makeatletter
\newcommand\figcaption{\def\@captype{figure}\caption}
\newcommand\tabcaption{\def\@captype{table}\caption}
\makeatother

\usepackage[capitalize]{cleveref}
\crefname{section}{Sec.}{Secs.}
\Crefname{section}{Section}{Sections}
\Crefname{table}{Table}{Tables}
\crefname{table}{Tab.}{Tabs.}

\title{No Time to Train:\\Empowering Non-Parametric Networks for Few-shot 3D Scene Segmentation}

\author{Xiangyang Zhu$^{*1,3}$, Renrui Zhang$^{*\dagger2,3}$, Bowei He$^{1}$, Ziyu Guo$^{2,3}$, Jiaming Liu$^{4}$ \\\vspace{0.2cm} Han Xiao$^{3}$, Chaoyou Fu$^{5}$, Hao Dong$^{4}$, Peng Gao$^{3}$\\
\normalsize{$^*$ Equal contribution}\quad  $\dagger$ Project leader\quad  \vspace{0.3cm}\\
  $^1$City University of Hong Kong \quad \vspace{0.07cm}
  $^2$The Chinese University of Hong Kong\\
  $^3$Shanghai AI Lab \quad
  $^4$Peking University\quad
  $^5$Tencent Youtu Lab\quad \\
\texttt{\{xiangyzhu6-c, boweihe2-c\}@my.cityu.edu.hk}\\
\texttt{\{zhangrenrui, gaopeng\}@pjlab.org.cn}\quad 
}

\begin{document}
\maketitle
\begin{abstract}
To reduce the reliance on large-scale datasets, recent works in 3D segmentation resort to few-shot learning. Current 3D few-shot segmentation methods first pre-train models on `seen' classes, and then evaluate their generalization performance on `unseen' classes. However, the prior pre-training stage not only introduces excessive time overhead but also incurs a significant domain gap on `unseen' classes. To tackle these issues, we propose a \textbf{N}on-parametric \textbf{N}etwork for few-shot 3D \textbf{Seg}mentation, \textbf{Seg-NN}, and its \textbf{P}arametric variant, \textbf{Seg-PN}. Without training, Seg-NN extracts dense representations by hand-crafted filters and achieves comparable performance to existing parametric models. Due to the elimination of pre-training, Seg-NN can alleviate the domain gap issue and save a substantial amount of time. Based on Seg-NN, Seg-PN only requires training a lightweight QUEry-Support Transferring (QUEST) module, which enhances the interaction between the support set and query set. Experiments suggest that Seg-PN outperforms previous state-of-the-art method by \textbf{+4.19\%} and \textbf{+7.71\%} mIoU on S3DIS and ScanNet datasets respectively, while reducing training time by  \textbf{-90\%}, indicating its effectiveness and efficiency. Code is available \href{https://github.com/yangyangyang127/Seg-NN}{here}.
\end{abstract}

\vspace{-0.5cm}
\section{Introduction}
\label{sec:intro}

Point cloud segmentation is an essential procedure in autonomous driving~\cite{krispel2020fuseseg, wang2020pillar}, robotics~\cite{lewandowski2019fast, ahmed2018edge}, and other computer vision applications~\cite{yue2018lidar, bello2020deep}. To achieve favorable segmentation, many learning-based methods have been proposed~\cite{li2022primitive3d, lai2022stratified, yang2023swin3d,  li2023transformer}. However, such algorithms require the construction of large-scale and accurately annotated datasets, which is expensive and time-consuming. 

\begin{figure}[t]
\subfloat[Existing Methods]{\includegraphics[width=0.45\textwidth]{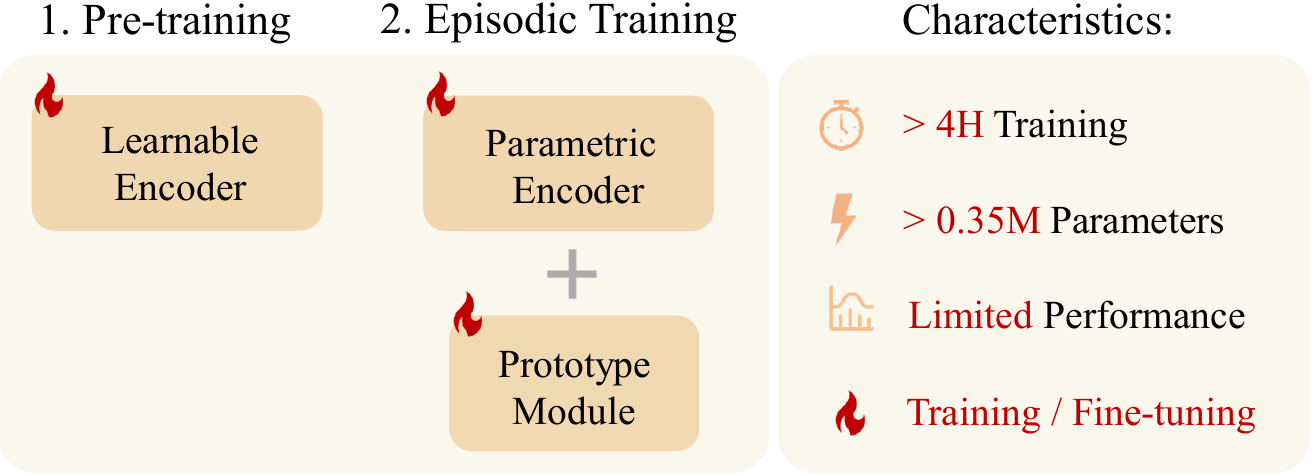}} \vspace{0.1cm} \\
\subfloat[Our \textbf{Seg-NN} or \textbf{Seg-PN}]{\includegraphics[width=0.45\textwidth]{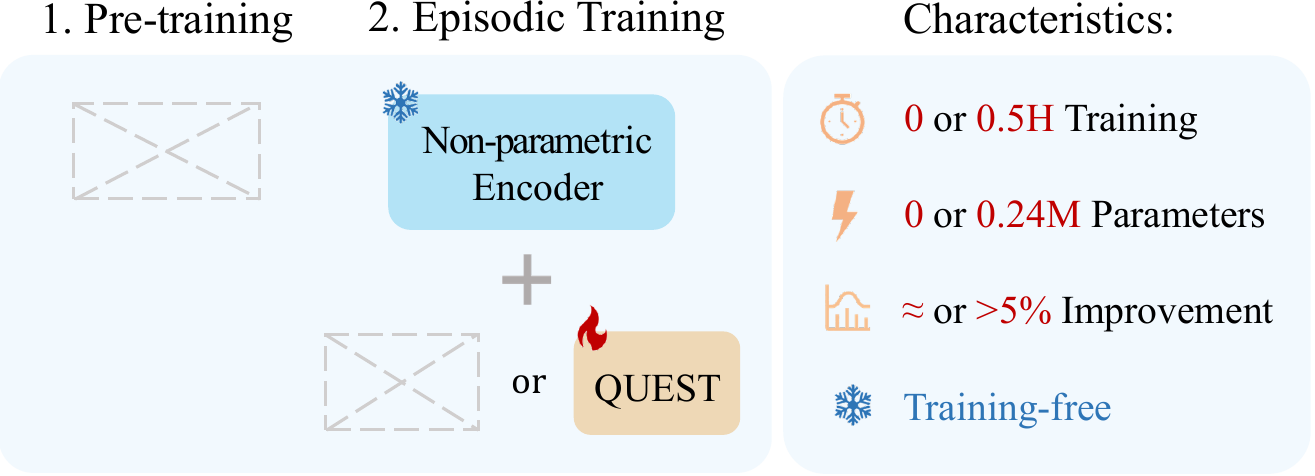}}
\vspace{-0.2cm}
\caption{\textbf{Comparison of Existing Methods and Our Approaches.} Our non-parametric Seg-NN contains no learnable parameters and thus discards both pre-training and episodic training stages with superior efficiency, and the parametric Seg-PN further improves the performance with a lightweight QUEST module.}
\label{fig:existing_comparison}
\vspace{-0.3cm}
\end{figure}

The data-hungry problem can be effectively mitigated by the few-shot learning strategy, which has garnered significant attention~\cite{he2023prototype}. 
From Fig. \ref{fig:existing_comparison} (a), existing 3D few-shot segmentation methods basically follow the meta-learning scheme to learn a 3D encoder and a prototype generation module. They mainly take three steps to solve the problem:
\vspace{-0.2cm}
\begin{enumerate}[label=\arabic*.]
  \item \textbf{Pre-training on `seen' classes} by supervised learning. Considering the lack of pre-trained models in the 3D field, this step trains a learnable 3D encoder, e.g., DGCNN~\cite{wang2019dynamic}, to obtain the ability to extract general 3D point cloud representations.
  \item \textbf{Episodic training on `seen' classes} to fit the query-support few-shot segmentation tasks. In this step, the pre-trained encoder is appended with a learnable prototype module and fine-tuned to extract discriminative prototypes from the support set, which are utilized to guide the semantic segmentation of the query set.
  \item \textbf{Testing on `unseen' classes} to evaluate the model. After episodic training, the model is evaluated on test episodes that contain unseen classes. The model is expected to segment new classes by the same query-support paradigms as the episodic training.
\end{enumerate}
\vspace{-0.1cm}
However, this pipeline encompasses two noteworthy issues: 1) the learnable encoder being pre-trained and fine-tuned on `seen' classes will inevitably introduce a domain gap when evaluated on `unseen' classes; 2) the complexity of the training process, including pre-training and episodic training, incurs substantial time and resource overhead.

\begin{figure}[t]
\centering
\subfloat[Performance Difference of Seen and Unseen Classes]{\includegraphics[width=0.22\textwidth]{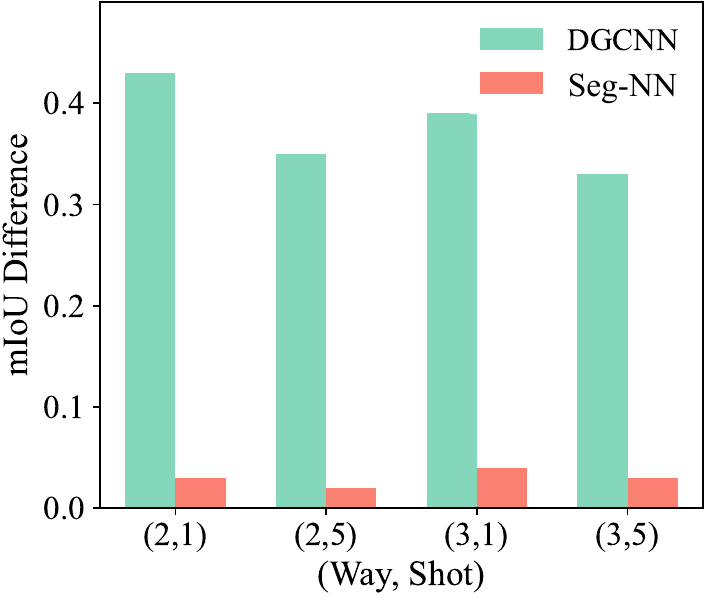}} 
\hspace{4pt}
\subfloat[KL Divergence of Logits between Support and Query Sets]{\includegraphics[width=0.22\textwidth]{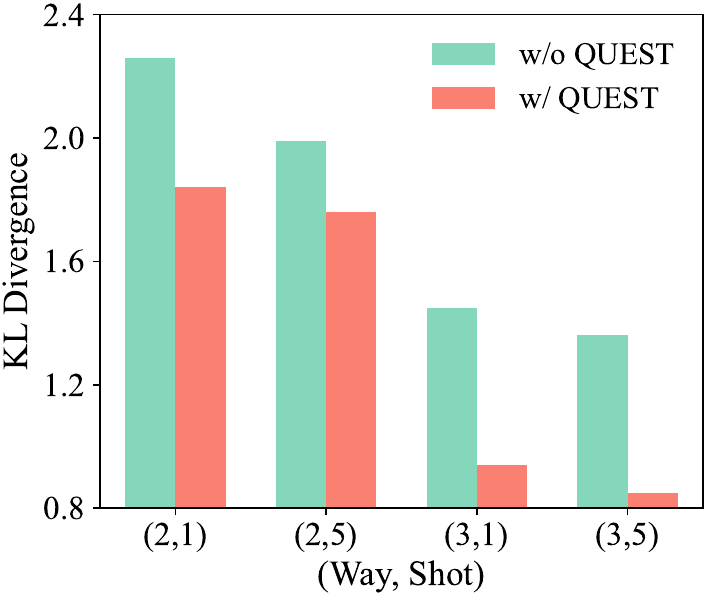}}
\vspace{-0.15cm}
\caption{\textbf{Alleviating Domain Gap by Seg-NN (a) and Prototype Bias by Seg-PN (b)} on S3DIS~\cite{armeni20163d}.
The horizontal axis presents the number of ways and shots in the form of (Way, Shot). 
}
\label{fig:efficacy_QUEST}
\vspace{-0.4cm}
\end{figure}

To address these issues, we extend the non-parametric network, Point-NN~\cite{zhang2023parameter}, to few-shot 3D scene segmentation tasks and propose a new model, \textbf{Seg-NN}, which is both efficient and effective. Seg-NN inherits the non-parametric encoder from Point-NN to encode 3D scenes but makes the following modifications: 1) We project the position and color information into a shared feature space and integrate them to obtain a comprehensive representation. 2) To reduce the noises and perturbations in natural 3D scenes, we sample the robust low frequencies and filter out the noisy high-frequency components. 
After encoding, we leverage the category prototypes to predict the segmentation masks for the query set by similarity matching. As shown in Fig. \ref{fig:existing_comparison} (b), Seg-NN discards all two stages of pre-training and episodic training and performs comparably to some existing parametric methods. Such a training-free property simplifies the few-shot training pipeline with minimal resource consumption and mitigates the domain gap caused by different training-test categories. As visualized in Fig. \ref{fig:efficacy_QUEST} (a), we observe that Seg-NN shows marginal performance difference between seen and unseen categories, while the widely adopted DGCNN~\cite{wang2019dynamic} encoder presents a much worse generalization ability due to cross-domain training and testing.

On top of this, we also propose a parametric variant, \textbf{Seg-PN}, to further boost the performance by efficient training. In detail, we adopt the non-parametric encoder of Seg-NN and append an additional parametric \textbf{QUE}ry-\textbf{S}upport \textbf{T}ransferring module, termed \textbf{QUEST}. QUEST enhances category prototypes in the support-set domain with the knowledge from the query-set domain, which suppresses prototype biases caused by the small few-shot support set. As shown in Fig. \ref{fig:efficacy_QUEST} (b), the reduced query-support distribution differences suggest that the prototypes are shifted to the query-set domain.
Seg-PN only learns the QUEST module and does not require pre-training just as Seg-NN, as shown in Fig. \ref{fig:existing_comparison} (b). Experiments show that Seg-PN achieves new state-of-the-art (SOTA) performance on both S3DIS~\cite{armeni20163d} and ScanNet~\cite{dai2017scannet} datasets, surpassing the second-best by \textbf{+4.19\%} and \textbf{+7.71\%}, respectively, while reducing the training time by over \textbf{-90\%}.

In summary, our contributions are as follows:

\begin{itemize}
    \item We introduce a non-parametric few-shot learning framework, Seg-NN, for 3D point cloud semantic segmentation, which can also serve as a basis to construct the parametric better-performed variant, Seg-PN.
    \item We design a new query-support interaction module, QUEST, in Seg-PN to adapt category prototypes by learning the affinity between support and query sets.
    \item  Comprehensive experiments are conducted to verify the efficiency and efficacy of our proposed method. We achieve SOTA results with the least parameters and a substantially simplified learning pipeline.
\end{itemize}


\begin{figure*}[!t]
\centering
\includegraphics[width=0.98\textwidth]{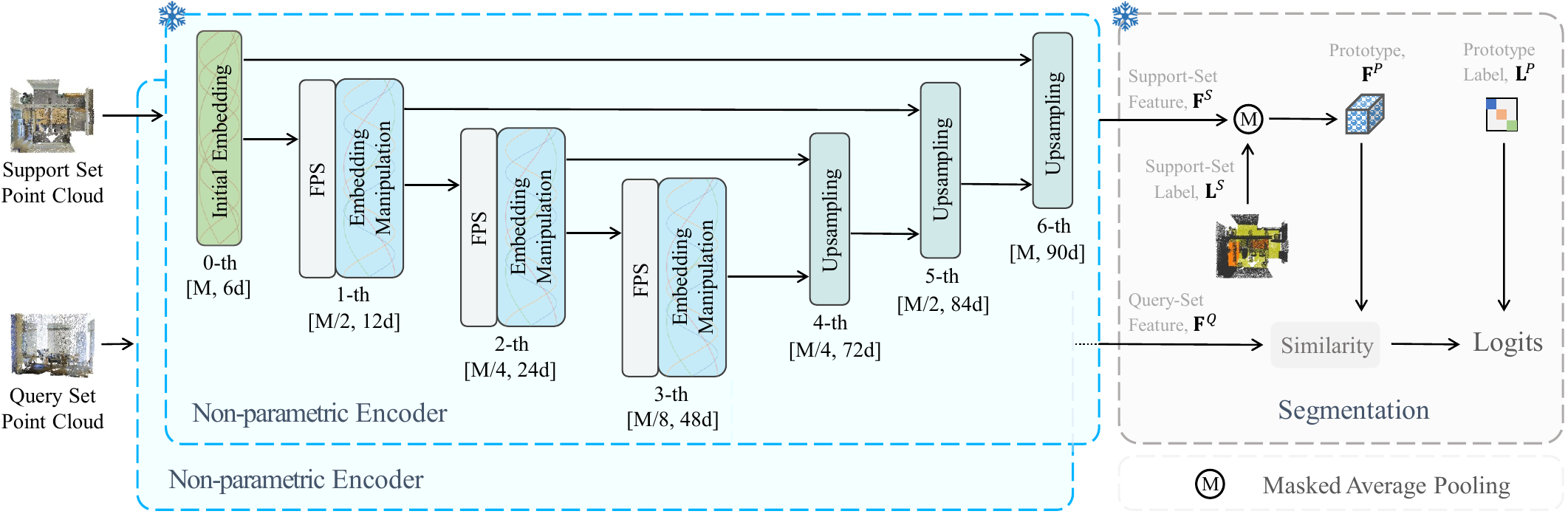}
\caption{\textbf{The Framework of the Non-parametric Seg-NN}. The encoder extracts support- and query-set features and the segmentation head segments the query set based on similarity. To facilitate illustration, we assume the encoder consists of three manipulation layers.}
\label{fig:framework}
\vspace{-0.35cm}
\end{figure*}

\section{Problem Definition}
\label{sec:problem_definition}

We first illustrate the task definition of few-shot 3D semantic segmentation. 
We follow previous works \cite{zhao2021few, he2023prototype} to adopt the popular episodic training/test paradigm \cite{vinyals2016matching} after the pre-training stage. Each episode is instantiated as an $N$-way $K$-shot task, which contains a support set and a query set. The support set comprises $N$ target classes, and each class corresponds to $K$ point clouds with point-level labels. The query set contains a fixed number of point clouds that need to be segmented.
Each episodic task aims to segment the query-set samples into $N$ target classes along with a `background' class based on the guidance of the support-set samples and embeddings.

To achieve this, we regard the semantic segmentation task as a point-level similarity-matching problem. We first utilize a feature encoder to extract the features of support-set point cloud samples and generate prototypes for all $N+1$ classes. Then, we adopt the same feature encoder to obtain the feature of every point in query samples and conduct similarity matching with the prototypes for point-level classification. In this way, the query-set point clouds can be segmented into $N+1$ semantic categories.

In the following two sections, we respectively illustrate the details of our proposed efficient frameworks, the non-parametric Seg-NN and parametric Seg-PN.

\section{Non-parametric Seg-NN}

The detailed structure of Seg-NN is shown in Fig. \ref{fig:framework}, which contains a U-Net \cite{ronneberger2015u} style training-free encoder and a similarity-based segmentation head. The encoder embeds the point cloud into high-dimensional representations, and the segmentation head conducts similarity matching to produce the final prediction. Overall, Seg-NN adopts a similar framework to Point-NN \cite{zhang2023parameter}, while we discard high-frequency noises to extract scene representations. The entire framework does not introduce any learnable parameters, thus deleting both pre-training and episodic training stages, differing from existing algorithms using DGCNN~\cite{wang2019dynamic} as the learning-required encoder. 
Following, we describe the details of the encoder and the segmentation head.

\subsection{Seg-NN Encoder}

Given a point cloud $\left \{ \mathbf{p}_{i} \right \}_{i=1}^{M}$ containing $M$ points, our goal is to encode each point into embedding space in a training-free manner. Taking a random point $\mathbf{p}=(x,y,z)$ as an example, we denote its RGB color as $\mathbf{c}=(r, g, b)$. The encoder aims to extract shape knowledge based on both position and color information. 

Firstly, we utilize trigonometric PEs to encode both positions and colors into high-dimension encodings as in Point-NN~\cite{zhang2023parameter}, which also map colors and positions into the same feature space.
Specifically, we embed $\mathbf{p}$ and $\mathbf{c}$ with $d$ frequencies $\mathbf{u}=[u_1, ..., u_d]$, denoted as $\operatorname{E}(\cdot)$: 
\begin{equation}
\setlength{\abovedisplayskip}{4pt}
\setlength{\belowdisplayskip}{3pt}
\begin{aligned}
\operatorname{E}(\mathbf{p};\mathbf{u}) = [ & \sin(2 \pi \mathbf{u} \mathbf{p}), \cos(2 \pi \mathbf{u} \mathbf{p})]  \quad \in \mathbb{R}^{6d}, \\
\operatorname{E}(\mathbf{c};\mathbf{u}) = [ & \sin(2 \pi \mathbf{u} \mathbf{c}),  \cos(2 \pi \mathbf{u} \mathbf{c})] \quad \in \mathbb{R}^{6d},
\end{aligned}
\label{equ:sine_embed}
\end{equation}
where $6d$ comes from the combination of 3 coordinates, $(x,y,z)$, with 2 functions, $\sin(\cdot)$ and $\cos(\cdot)$. All frequency components in $\mathbf{u}$ adhere to a log-linear form, $u_{i}=\theta^{i/d}, i=1,..., d$, where $\theta$ and $d$ are hyperparameters.
Fig. \ref{fig:example_encoding} (a)(b) presents two encoding examples. The trigonometric PEs can not only encode the absolute point positions but also reveal the relative spatial relations between different points. For two points, $\mathbf{p_i}$ and $\mathbf{p_j}$, their embeddings $\operatorname{E}(\mathbf{p_i}; \mathbf{u})$ and $\operatorname{E}(\mathbf{p_j}; \mathbf{u})$ represent their absolute positional information. Then, the relative relation, $\mathbf{p_i} - \mathbf{p_j}$, can also be preserved by the dot production as $\operatorname{E}(\mathbf{p_i}; \mathbf{u}) \operatorname{E}(\mathbf{p_j}; \mathbf{u})^T$\ =\ $\sum_{m=1}^{d}\cos\big((\mathbf{p_i}-\mathbf{p_j})u_m\big)$. Therefore, by such spatial-aware encoding, we effectively vectorize the point clouds.

We designate this as the 0-th layer and merge the position and color vector for a comprehensive embedding, $\mathbf{f}^{0}$,
\begin{equation}
\setlength{\abovedisplayskip}{4pt}
\setlength{\belowdisplayskip}{4pt}
\begin{aligned}
\mathbf{f}^{0} = \operatorname{E}(\mathbf{p};\mathbf{u}) + \operatorname{E}(\mathbf{c};\mathbf{u}) \quad \in \mathbb{R}^{6d}.
\end{aligned}
\label{equ:initial_integrate}
\end{equation}
This trigonometric initial embedding can effectively represent position and color information. After that, we adopt two types of layers to further extract deep 3D embeddings as shown in Fig. \ref{fig:framework}: stacked \textbf{\textit{Embedding Manipulation}} layers extract hierarchical embeddings by groups of hand-crafted filters; \textbf{\textit{Upsampling}} layers integrate and upsample hierarchical embeddings to obtain the final point-level features.

\vspace{-0.2cm}
\subsubsection{ Embedding Manipulation} 
As presented in Fig. \ref{fig:framework}, we stack embedding manipulation layers to get deep representations, which encode local shapes at different scales. Before each layer, we downsample the point cloud by half to increase the receptive field with Farthest Point Sampling (FPS).    
To obtain local embeddings, we denote point $\mathbf{p}$ as a local center and consider its neighborhood $\mathcal{N}_p$ searched by the $k$-Nearest Neighbor ($k$-NN) algorithm. We concatenate the neighbor point feature with the center feature along the channel dimension,
\begin{equation}
\setlength{\abovedisplayskip}{4pt}
\setlength{\belowdisplayskip}{3pt}
\begin{aligned}
\hat{\mathbf{f}}_{j}^{l} = \operatorname{Concat}(\mathbf{f}^{l-1}, \mathbf{f}_{j}^{l-1}), \quad j \in \mathcal{N}_p,
\end{aligned}
\label{equ:local_concat}
\end{equation}
where the subscript $j$ represents the $j$-th point in $\mathbf{p}$'s neighborhood $\mathcal{N}_p$, and $l$ denotes the $l$-th layer as in Fig. \ref{fig:framework}. After concatenation, the expanded embedding $\hat{\mathbf{f}}_{j}^{l} \in \mathbb{R}^{2^{l} \times 6d}$  incorporate both center and neighbor information. The dimensionality is increased as $2^{l}\times 6d$ since we conduct Eq. \ref{equ:local_concat} at every manipulation layer, each of which doubles the channel dimension.
Then, to reveal local patterns, we design a group of filters based on previous efforts.

\begin{figure}[t!]
\centering
\hspace{-5pt}\subfloat[Point Encoding Example 1]{\includegraphics[width=0.23\textwidth]{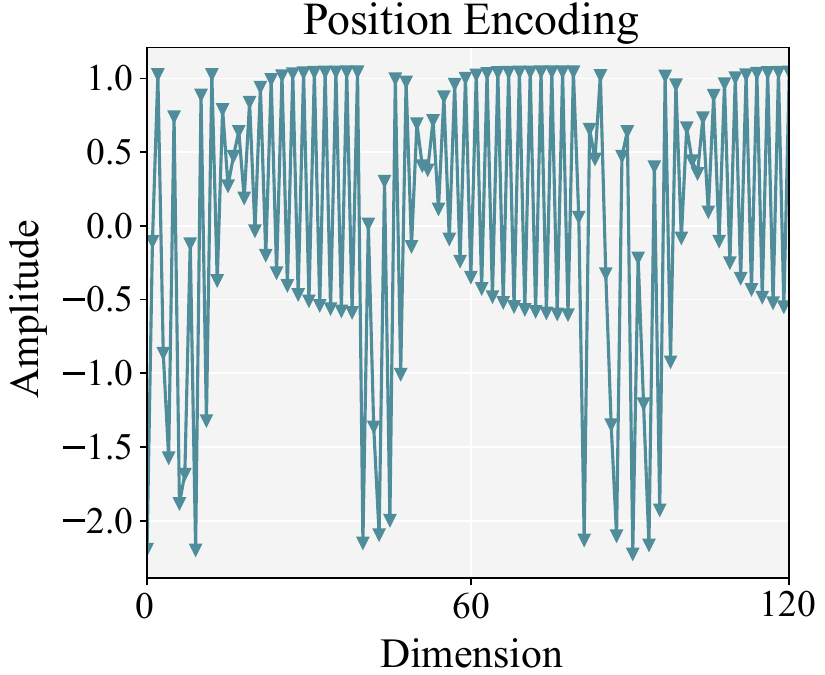}} 
\hspace{2pt}
\subfloat[Point Encoding Example 2]{\includegraphics[width=0.23\textwidth]{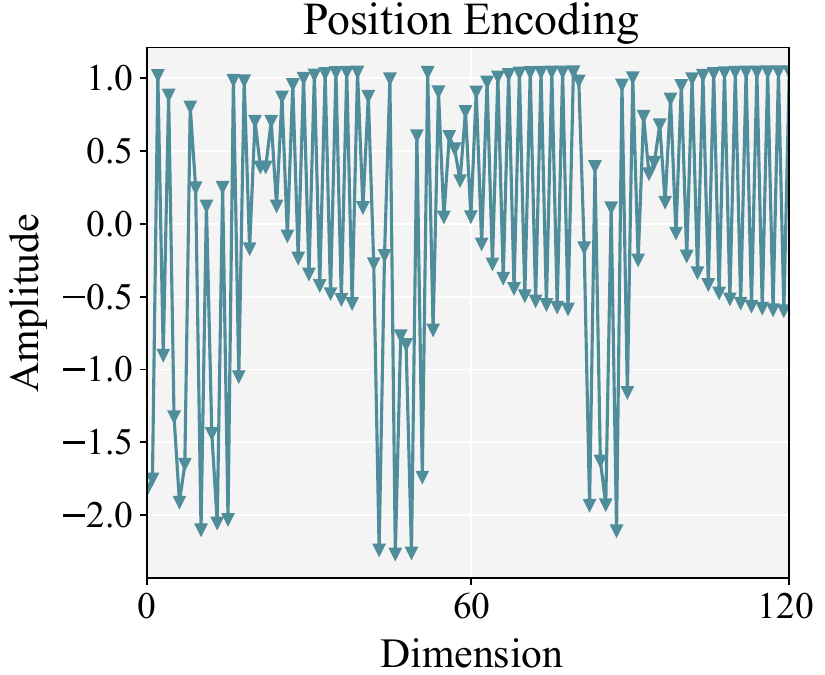}}\vspace{4pt}\\
\hspace{-5pt}\subfloat[Frequency Spectrum]{\includegraphics[width=0.23\textwidth]{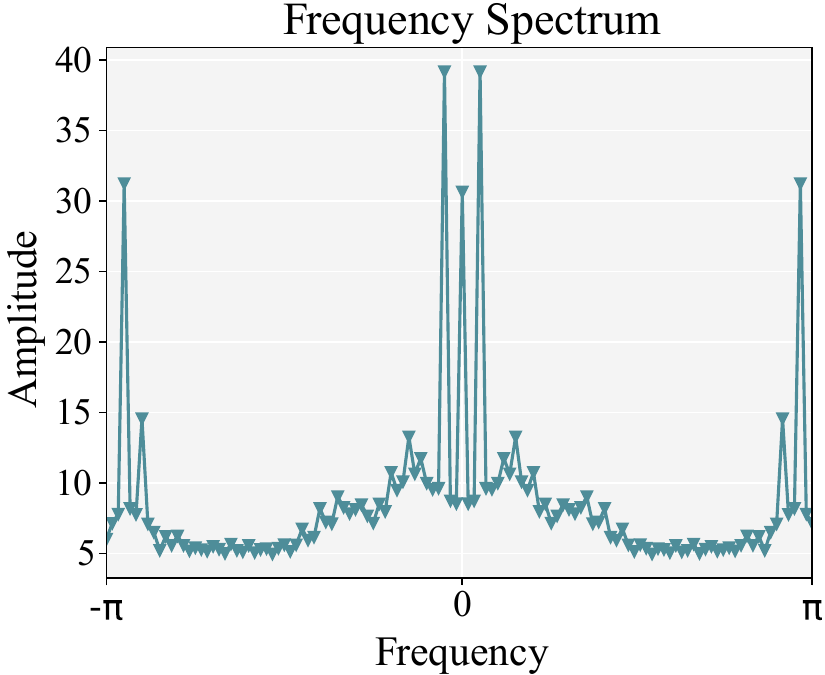}}
\hspace{2pt}
\subfloat[Distribution of Sampling]{\includegraphics[width=0.23\textwidth]{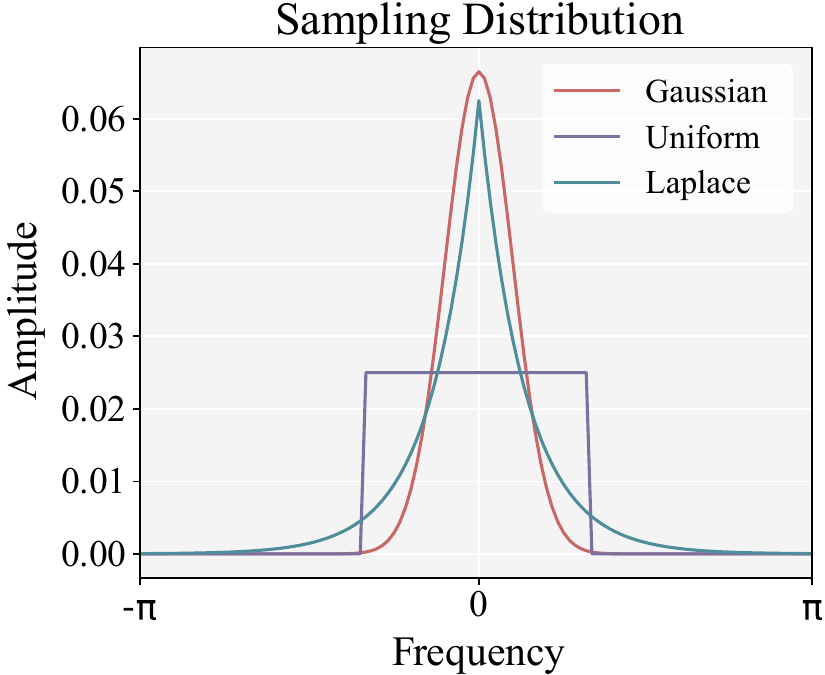}}
\vspace{-0.15cm}
\caption{\textbf{Examples of Position Encodings and Frequencies,} where we set $d=20$. (a) and (b) are examples of initial encodings. (c) is the average frequency spectrum over all points' encodings of a point cloud. (d) is the distribution of sampled frequencies.}
\label{fig:example_encoding}
\vspace{-0.4cm}
\end{figure}

Existing works, \eg,  PointNet++~\cite{qi2017pointnet++}, DGCNN~\cite{wang2019dynamic}, and Point-MLP~\cite{ma2022rethinking}, efficiently extract features via multiple learning-required linear projections, which can be regarded as diverse filters from a frequency perspective. Inspired by this, we seek to manually design the filters. From Fig. \ref{fig:example_encoding} (a)(b), we can easily observe that the initial point encodings are band-limited signals. As shown in Fig. \ref{fig:example_encoding} (c), embedding frequencies are mainly distributed in the low- and high-frequency ranges. In this work, we aim to encode natural 3D scene points, which generally contain noises and perturbations. The large sharp noises contained in high frequencies may lead to significant differences in embeddings from clean points. Additionally, research has proved that neural networks tend to prioritize the learning of low-frequency information~\cite{zhang2019explicitizing, luo2019theory}, which indicates that low frequencies are discriminative and robust. Therefore, we rely on low-frequency bands to extract features and filter out high frequencies to further refine the scene representation.

We randomly sample low frequencies to construct a linear projection. As illustrated in Fig. \ref{fig:example_encoding} (d), we can follow Gaussian, uniform, Laplace, or other distributions to sample the frequencies. We denote them as $\mathbf{v} \in \mathbb{R}^{2^{l} \times 6d}$. Then, the projection weight is $\mathbf{W}^{l} = [\cos(2 \pi \mathbf{v} \mathbf{k})] \in \mathbb{R}^{(2^{l} \times 6d) \times (2^{l} \times 6d)}$, where $\mathbf{k}=[1, ..., 2^{l} \times 6d]$. We integrate the relative position $\Delta \mathbf{p}_{j}$ from neighbor point $j$ to the center point and color information for a comprehensive embedding. Thus, point $j$'s embedding can be calculated via 
\begin{equation}
\setlength{\abovedisplayskip}{3pt}
\setlength{\belowdisplayskip}{4pt}
\begin{aligned}
    &\mathbf{f}^{l}_{j} = \mathbf{W}^{l} \cdot \big(\hat{\mathbf{f}}_{j}^{l} + \operatorname{E}(\Delta \mathbf{p}_{j}; \mathbf{u}) + \operatorname{E}( \mathbf{c}_{j}; \mathbf{u})\big), 
\end{aligned}
\label{equ:add_multiply}
\end{equation}
where $\mathbf{u}$ are log-linear frequencies similar to the initial encoding and $\mathbf{f}^{l}_{j} \in \mathbb{R}^{2^{l} \times 6d}$. 
Finally, we use maximum pooling to compress the neighborhood information into the central point and use $\mathbf{f}^{l}\in \mathbb{R}^{2^{l} \times 6d}$ to represent the embedding of point $\mathbf{p}$ in the $l$-th layer.
We acquire hierarchical features $\{ \mathbf{f}^{l}\}|_{l}$ from the manipulation layers and subsequently feed them into upsampling layers.

\vspace{-0.3cm}
\subsubsection{Upsampling}
\vspace{-0.1cm}
After all manipulation layers, we adopt the upsampling operation in Point-NN \cite{zhang2023parameter} to progressively upsample the hierarchical features to the input point number, as in Fig. \ref{fig:framework}. Specifically, we first interpolate the central point embedding via the weighted sum of neighbor point embeddings. Then, we concatenate the embedding of manipulation layers and upsampling layers in the channel dimension as the output. 

After the upsampling layers, we obtain the final representation for each point, denoted as $\mathbf{f} \in \mathbb{R}^{D}, D =(2^{0}+...+2^{3})\times 6d$. In this way, the encoder produces point-level embeddings of the input point cloud, denoted as $\mathbf{F}=\{ \mathbf{f}_{m} \} ^{M}_{m=1}$, $\mathbf{F} \in \mathbb{R}^{M \times D}$, where $M$ is point number.

\vspace{-0.4cm}
\paragraph{} By stacking these layers, the encoder can encode scene points without learnable parameters, so it eliminates tedious pre-training and episodic training. This advantage allows Seg-NN to save significant time and resources and alleviate domain gaps caused by disparate training and test classes.

\subsection{Similarity-based Segmentation}

We utilize the non-parametric encoder to respectively extract support sample features, denoted as $\mathbf{F}^{S} \in \mathbb{R}^{M \times D}$, and query sample features, $\mathbf{F}^{Q}\in \mathbb{R}^{M \times D}$. We also denote the support sample labels as $\mathbf{L}^{S} \in \mathbb{R}^{M}$. Then, we conduct a simple similarity-based segmentation~\cite{zhang2022nearest,zhang2022point}.
We use masked average pooling \cite{he2023prototype} to produce the prototypes of $N+1$ classes, denoted as $\mathbf{F}^{P} \in \mathbb{R}^{ (N+1)\times D}$.
Then, the cosine similarity between normalized $\mathbf{F}^{Q}$ and $\mathbf{F}^{P}$ is, 
\begin{equation}
\setlength{\abovedisplayskip}{3pt}
\setlength{\belowdisplayskip}{3pt}
\begin{aligned}
    \mathbf{S}_{cos} = \mathbf{F}^{Q}\mathbf{F}^{P}{}^{\top} \in \mathbb{R}^{M \times (N+1)},
\end{aligned}
\end{equation}
which represents the similarity between each point in the query set and $N+1$ prototypes.
Finally, we weight and integrate the one-hot labels of the prototypes, $\mathbf{L}^{P} \in \mathbb{R}^{(N+1) \times (N+1)}$, to achieve the final prediction, 
\begin{equation}
\setlength{\abovedisplayskip}{3pt}
\setlength{\belowdisplayskip}{3pt}
\begin{aligned}
    \mathrm{logits} = \varphi(\mathbf{S}_{cos} \mathbf{L}^{P})\ \in \mathbb{R}^{M \times (N+1)},
\end{aligned}
\end{equation}
where $\varphi(x)=\exp(-\gamma(1-x))$ acts as an activation function and $\gamma$ is a scaling factor~\cite{zhang2021tip,zhang2023prompt}.
For $N$-way-$K$-shot tasks, we produce $K$ prototypes for each category. 
By this segmentation head, the entire Seg-NN framework can be purely training-free, thus achieving superior efficiency.

\section{Parametric Seg-PN}

To further achieve better performance, we propose a parametric version, Seg-PN. 
Seg-PN inherits the non-parametric 3D encoder of Seg-NN to encode point clouds and only introduces a learnable lightweight segmentation head, QUEST. In contrast, the parametric version of Point-NN, Point-PN~\cite{zhang2023parameter}, inserts learnable layers into Point-NN's encoder and requires time-consuming training. 

\begin{figure}[t]
\centering
\includegraphics[width=0.48\textwidth]{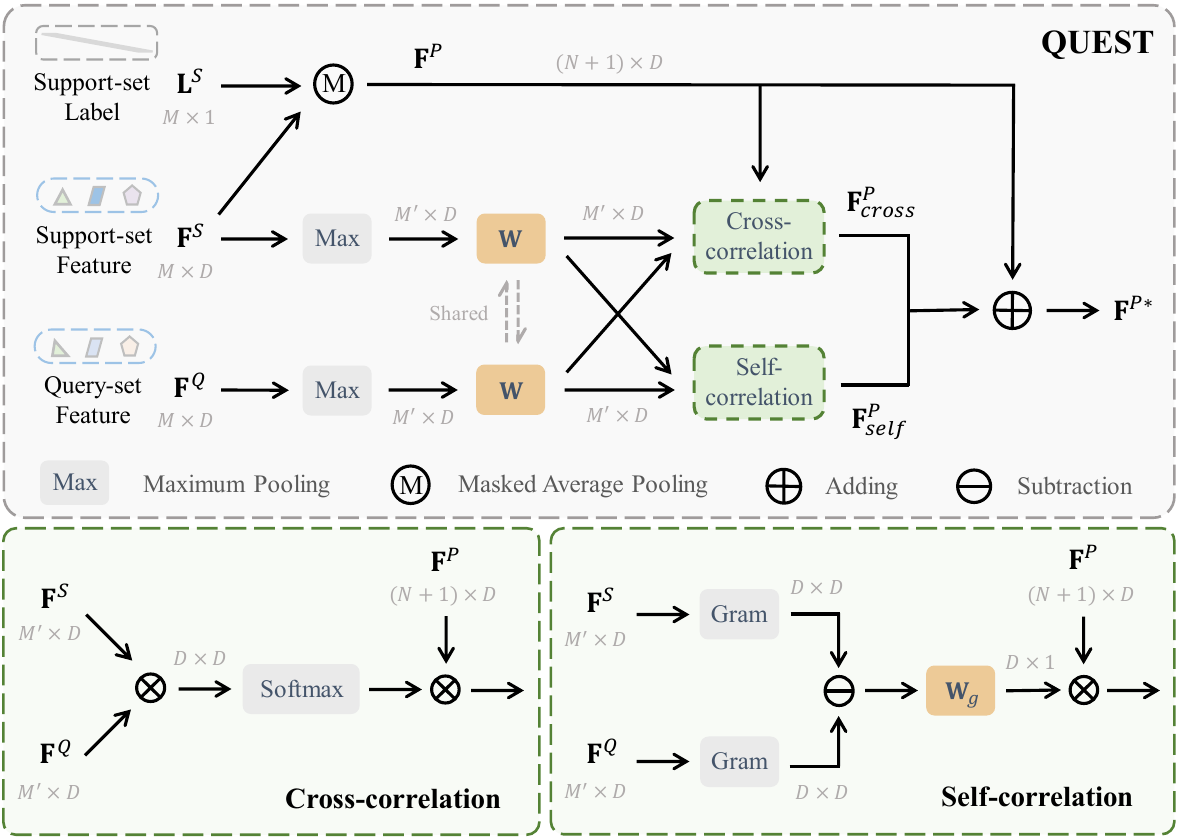}
\vspace{-0.5cm}
\caption{\textbf{Details of QUEST in Seg-PN.} QUEST finally outputs adjusted prototypes $\mathbf{F}^{P}{}^{*}$.}
\label{fig:SegPN}
\vspace{-0.4cm}
\end{figure}

An obvious problem in few-shot learning is that the small-scale support set might fail to represent the true distribution of each category, leading to biased prototypical learning. We propose a Query-Support Transferring module, QUEST, to mitigate this problem and transfer the prototypes from the support-set to the query-set domain. QUEST directly follows the encoder and adjusts the prototypes based on query-support interaction.
The detailed structure of QUEST is shown in Fig. \ref{fig:SegPN}. We first conduct local maximum pooling along the point dimension to obtain statistics of each feature channel in the support-set and query-set features.
Then, we leverage a shared projection operation, $\mathbf{W}$, to refine the statistics of each channel. This step can be denoted as
\begin{equation}
\setlength{\abovedisplayskip}{3pt}
\setlength{\belowdisplayskip}{3pt}
\begin{aligned}
    &\mathbf{F}^{S} = \mathbf{W} \cdot \operatorname{MaxPool} (\mathbf{F}^{S}) \quad \in \mathbb{R}^{M' \times D}, \\
    &\mathbf{F}^{Q} = \mathbf{W} \cdot \operatorname{MaxPool} (\mathbf{F}^{Q}) \quad \in \mathbb{R}^{M' \times D},
\end{aligned}
\label{equ:QUEST_maxpool}
\end{equation}
where the kernel size and stride of the pooling operation are hyperparameters. Eq. \ref{equ:QUEST_maxpool} produces $M'$ statistics from the $M$ points to manifest the distribution of each feature channel. We use the same way to obtain $\mathbf{F}^{P}$ as in Seg-NN. On top of this, QUEST bridges the support and query data under self-correlation and cross-correlation schemes.

\vspace{-0.2cm}
\paragraph{Cross-correlation.} We investigate the cross-correlation of the geometric structures between the support set and query set, which is utilized to adjust the category prototypes $\mathbf{F}^{P}$. Specifically, the cross-correlation between $\mathbf{F}^{S}$ and $\mathbf{F}^{Q}$ is calculated as their inner product, 
\begin{equation}
\setlength{\abovedisplayskip}{4pt}
\setlength{\belowdisplayskip}{4pt}
\begin{aligned}
    \mathbf{C}_{cross} = {\mathbf{F}^{Q}}{}^\top \mathbf{F}^{S} \quad \in \mathbb{R}^{D \times D},
\end{aligned}
\end{equation}
where the diagonal elements of $\mathbf{C}_{cross}$ are the correlation between the geometric structures of the support sample and query sample, while the non-diagonal elements reflect the cross-correlation between different channels of the support-set and query-set features. Next, we modulate $\mathbf{C}_{cross}$ using the softmax function and use it to adjust the corresponding channels of $\mathbf{F}^{P}$ by matrix multiplication,
\begin{equation}
\setlength{\abovedisplayskip}{4pt}
\setlength{\belowdisplayskip}{4pt}
\begin{aligned}
\mathbf{F}^{P}_{cross} = \operatorname{Softmax}(\mathbf{C}_{cross}) \cdot \mathbf{F}^{P}{}^\top .
\label{eq:get_FPcross}
\end{aligned}
\end{equation}
From Eq. \ref{eq:get_FPcross}, we obtain the adjusted category prototypes via query-support cross-correlation, which learns to transfer the prototypes $\mathbf{F}^P$ to the query-set domain. Compared to the cross-attention module in \cite{he2023prototype} and the bias rectification operation in \cite{zhu2023cross}, our cross-correlation scheme can better capture domain bias between the support set and query set.

\begin{table*}[t!]
\centering
\begin{adjustbox}{width=0.99\linewidth}
\begin{tabular}{l|c||ccc|ccc||ccc|ccc}
\toprule
\multirow{3}{*}{\textbf{Method}} & \multirow{3}{*}{\textbf{Param.}}
	& \multicolumn{6}{c||}{\textbf{Two-way}} & \multicolumn{6}{c}{\textbf{Three-way}}\\
	\cline{3-14}
	& & \multicolumn{3}{c|}{\textbf{One-shot}} & \multicolumn{3}{c||}{\textbf{Five-shot}} & 
	\multicolumn{3}{c|}{\textbf{One-shot}} & \multicolumn{3}{c}{\textbf{Five-shot}} \\
	\cline{3-14}
	& & $S_0$ & $S_1$ & $Avg$ & $S_0$ & $S_1$ & $Avg$ & $S_0$ & $S_1$ & $Avg$ & $S_0$ & $S_1$ & $Avg$ \\
	\midrule

    \rowcolor{purple!5} Point-NN~\cite{zhang2023parameter} & 0.00~M & 42.12 & 42.62 & 42.37 & 51.91 & 49.35 & 50.63 & 38.00 & 36.21 & 37.10 & 45.91 & 43.44 & 44.67\\

    \rowcolor{purple!5} \textbf{Seg-NN} & 0.00~M & {49.45} & {49.60} & 49.53 & {59.40} & {61.48} & 60.44 & 39.06 & 40.10 & 39.58 & 50.14 & 51.33 & 50.74 \\

    \rowcolor{purple!5} \textit{\small{Improvement}} & - & \textcolor{blue}{\small{+7.33}} & \textcolor{blue}{\small{+6.98}} & \textcolor{blue}{\small{+7.16}} & \textcolor{blue}{\small{+7.49}} & \textcolor{blue}{\small{+12.13}} & \textcolor{blue}{\small{+9.81}} & \textcolor{blue}{\small{+1.06}} & \textcolor{blue}{\small{+3.89}} & \textcolor{blue}{\small{+2.48}} & \textcolor{blue}{\small{+4.23}} & \textcolor{blue}{\small{+7.89}} & \textcolor{blue}{\small{+6.07}} \\
	\midrule
    DGCNN~\cite{wang2019dynamic} & {0.62~M} & 36.34 & 38.79 & 37.57 & 56.49 & 56.99 & 56.74 & 30.05 & 32.19 & 31.12 & 46.88 & 47.57 & 47.23 \\
    ProtoNet~\cite{garcia2017few} & 0.27~M & 48.39 & 49.98 & 49.19 & 57.34 & 63.22 & 60.28 & 40.81 & 45.07 & 42.94 & 49.05 & 53.42 & 51.24 \\
    
    MPTI~\cite{zhao2021few} & 0.29~M & 52.27 & 51.48 & 51.88 & 58.93 & 60.56 & 59.75 & 44.27 & 46.92 & 45.60 & 51.74 & 48.57 & 50.16 \\ 
    
    AttMPTI~\cite{zhao2021few} & 0.37~M & 53.77 & 55.94 & 54.86 & 61.67 & 67.02 & 64.35 & 45.18 & 49.27 & 47.23 & 54.92 & 56.79 & 55.86 \\ 
    BFG~\cite{mao2022bidirectional} & - & 55.60 & 55.98 & 55.79 & 63.71 & 66.62 & 65.17 & 46.18 & 48.36 & 47.27 & 55.05 & 57.80 & 56.43 \\ 
 
    2CBR~\cite{zhu2023cross} & 0.35~M & 55.89 & 61.99 & 58.94 & 63.55 & 67.51 & 65.53 & 46.51 & 53.91 & 50.21 & 55.51 & 58.07 & 56.79 \\ 
	PAP3D~\cite{he2023prototype} & 2.45~M & 59.45 & 66.08 & 62.76 & 65.40 & 70.30 & 67.85 & 48.99 & 56.57 & 52.78 & 61.27 & 60.81 & 61.04 \\
    \midrule
    
    \textbf{Seg-PN} & \textbf{0.24~M} & \textbf{64.84} & \textbf{67.98} & \textbf{66.41} & \textbf{67.63} & \textbf{71.48} & \textbf{69.36} & \textbf{60.12} & \textbf{63.22} & \textbf{61.67} & \textbf{62.58} & \textbf{64.53} & \textbf{63.56}  \\
    \textit{\small{Improvement}} & - & \textcolor{blue}{\small{+5.39}} & \textcolor{blue}{\small{+1.90}} & \textcolor{blue}{\small{+3.65}} & \textcolor{blue}{\small{+2.23}} & \textcolor{blue}{\small{+1.18}} & \textcolor{blue}{\small{+1.71}} & \textcolor{blue}{\small{+11.13}} & \textcolor{blue}{\small{+6.65}} & \textcolor{blue}{\small{+8.89}} & \textcolor{blue}{\small{+1.31}} & \textcolor{blue}{\small{+3.72}} & \textcolor{blue}{\small{+2.52}} \\
    
    \bottomrule
\end{tabular}
\end{adjustbox}
\caption{\textbf{Few-shot Results (\%) on S3DIS.} $S_i$ denotes the split $i$ is used for testing, and $Avg$ is their average mIoU. The shaded rows represent non-parametric methods. `Param.' represents the total number of learnable parameters of each method.}
\label{table:s3dis_iou}
\vspace{-0.3cm}
\end{table*}

\vspace{-0.2cm}
\paragraph{Self-correlation.} 
It is obvious that different feature channels extracted from our hand-crafted filters are not independent and certain correlations between channels exist. Further, this type of correlation exhibits differences between support-set and query-set domains. Therefore, we examine the difference in self-correlation between $\mathbf{F}^{S}$ and $\mathbf{F}^{Q}$, which measures the domain gap between the support set and query set. Generally, the Gram matrix of the feature vector is leveraged to represent the self-correlation~\cite{jing2019neural, lang2022learning}. Denoted as $\mathbf{G}^{S}$ and $\mathbf{G}^{Q}$, the Gram matrices of the support-set and query-set features are calculated as
\begin{equation}
\setlength{\abovedisplayskip}{4pt}
\setlength{\belowdisplayskip}{4pt}
\begin{aligned}
    \mathbf{G}^{S} = {\mathbf{F}^{S}}{}^\top \mathbf{F}^{S}, \quad
    \mathbf{G}^{Q} = {\mathbf{F}^{Q}}{}^\top \mathbf{F}^{Q}  \quad \in \mathbb{R}^{D \times D},
\end{aligned}
\end{equation}
both of which are symmetric matrices.
The diagonal elements of the Gram matrix reflect the characteristics of each channel itself, while the non-diagonal elements reflect the dependence between different channels. In addition, the difference between $\mathbf{G}^{S}$ and $\mathbf{G}^{Q}$ measures the domain gap between support and query sets. We extract this domain gap with a linear projection, $\mathbf{W}_{g} \in \mathbb{R}^{D \times 1}$, and use it to rectify the channels of $\mathbf{F}^{P}$:
\begin{equation}
\setlength{\abovedisplayskip}{4pt}
\setlength{\belowdisplayskip}{4pt}
\begin{aligned}
    \mathbf{F}^{P}_{self} = \mathbf{F}^{P}\operatorname{diag}((\mathbf{G}^{Q} - \mathbf{G}^{S})\mathbf{W}_{g}),
\end{aligned}
\end{equation}
where $\operatorname{diag}(\cdot)$ is diagonalization and $\mathbf{F}^{P}_{self}$ is the rectified prototypes via self-correlation. In this way, we shift the prototype to the query-set domain.

\vspace{-0.2cm}
\paragraph{} We integrate the adjusted prototypes via self- and cross-correlation and original prototypes to obtain the final prototypes $\mathbf{F}^{P}{}^{*}= \mathbf{F}^{P} + \mathbf{F}^{P}_{self} + \mathbf{F}^{P}_{cross}$. For $N$-way-$K$-shot problems, we average all $K$-shot adjusted prototypes. 

\vspace{-0.2cm}
\paragraph{} After QUEST, we utilize the same similarity-matching scheme as Seg-NN to segment the query set. As the encoder is training-free, we do not need pre-training and only require episodic training to learn the QUEST module. During training, we adopt cross-entropy loss to optimize QUEST.

\section{Experiments}
\vspace{-0.1cm}
In this section, we first introduce the datasets and implementation details. Then we report the experimental results in comparison with existing approaches. At last, we present ablation studies to verify the effectiveness.

\subsection{Experimental Details}

\paragraph{Datasets.} 
We validate our method using two public 3D datasets, S3DIS~\cite{armeni20163d} and ScanNet~\cite{dai2017scannet}. 
Due to the large scale of original scenes, we adopt the data pre-processing in \cite{zhao2021few, he2023prototype} and partition them into small blocks. Then S3DIS and ScanNet contain 7,547 and 36,350 blocks, respectively. $M=2048$ points are randomly sampled from each block. For each dataset, we generate a training class set $C_{train}$ and a test class set $C_{test}$ that have no overlap. We use $C_{train}$ for episodic training and $C_{test}$ for testing, performing cross-validation for each dataset. For $N$-way-$K$-shot test episodes, we iterate over all combinations of $N$ classes from $C_{test}$, sampling 100 episodes for each combination.

\vspace{-0.3cm}
\paragraph{Basic Settings.} 
For few-shot settings, we experiment under 2/3-way-1/5-shot settings respectively, following \cite{zhao2021few, mao2022bidirectional, he2023prototype}. For performance, we adopt the mean Intersection over Union (mIoU) as evaluation criteria. mIoU is computed by averaging the IoU scores across all unseen classes $C_{test}$. In Seg-NN, we set the frequency number in the initial embedding layer to $d=20$ and sample log-linear spaced frequencies $\mathbf{u}$ with $\theta=30$. In embedding manipulation layers, we sample frequencies $\mathbf{v}$ from a Gaussian distribution with variance 1. In the segmentation head, we set the scale factor $\gamma$ to 1000. The non-parametric encoder contains totally three manipulation layers. We provide more detailed settings in supplementary materials. 

\subsection{Analysis}
\paragraph{Baselines} To evaluate our method, we compare it with two types of methods. First, we consider parametric 2D/3D few-shot segmentation methods, including DGCNN~\cite{wang2019dynamic, shaban2017one}, ProtoNet~\cite{garcia2017few}, MTPI~\cite{zhao2021few}, AttMPTI~\cite{zhao2021few}, BFG~\cite{mao2022bidirectional}, 2CBR~\cite{zhu2023cross}, and PAP3D~\cite{he2023prototype}. Second, we re-implement the non-parametric model, Point-NN~\cite{zhang2023parameter}, under our settings. We report the results in Tab. \ref{table:s3dis_iou} and \ref{table:scannet_iou}.

\vspace{-0.3cm}
\paragraph{Performance.} For non-parametric Seg-NN, we compare it to Point-NN and observe significant improvement on the S3DIS dataset. In addition, Seg-NN even outperforms some parametric methods, such as DGCNN and ProtoNet without any training. For parametric Seg-PN, its results significantly outperform previous SOTA mIoU across all four few-shot tasks on 2 datasets. We achieve an average improvement of \textbf{+4.19\%} and \textbf{+7.71\%} across four tasks on the S3DIS and ScanNet datasets, respectively, demonstrating that our method can better alleviate domain gaps between seen and unseen classes. This also indicates the non-parametric encoder can extract discriminative and general knowledge for 3D shapes. From the perspective of parameter number, we only utilize 0.24M parameters, the least among existing methods and \textbf{-90\%} less than PAP3D with better performance.

\vspace{-0.3cm}
\paragraph{Efficiency.} In Tab. \ref{table:efficiency_comparison}, we compare the training time with existing works. 
Seg-NN achieves few-shot segmentation with minimal resource and time consumption as training is not required. For Seg-PN, we require only episodic training without pre-training, greatly reducing training time by more than \textbf{-90\%} compared to existing methods. Both Seg-NN and Seg-PN efficiently simplify the few-shot pipeline by discarding the pre-training step.

\begin{table*}[t]
\centering
\begin{adjustbox}{width=0.99\linewidth}
\begin{tabular}{l|c||ccc|ccc||ccc|ccc}
\toprule
\multirow{3}{*}{\textbf{Method}}
	& \multirow{3}{*}{\textbf{Param.}} & \multicolumn{6}{c||}{\textbf{Two-way}} & \multicolumn{6}{c}{\textbf{Three-way}}  \\
	\cline{3-14}
	& & \multicolumn{3}{c|}{\textbf{One-shot}} & \multicolumn{3}{c||}{\textbf{Five-shot}} &
	\multicolumn{3}{c|}{\textbf{One-shot}} & \multicolumn{3}{c}{\textbf{Five-shot}} \\
	\cline{3-14}
	& & $S_0$ & $S_1$ & $Avg$ & $S_0$ & $S_1$ & $Avg$ & $S_0$ & $S_1$ & $Avg$ & $S_0$ & $S_1$ & $Avg$ \\
    \midrule

    \rowcolor{purple!5} Point-NN~\cite{zhang2023parameter} & 0.00~M & 28.85 & 31.56 & 30.21 & 34.82 & 32.87 & 33.85 & 21.24 & 17.91 & 19.58 & 26.42 & 23.98 & 25.20 \\
    
    \rowcolor{purple!5} \textbf{Seg-NN} & 0.00~M & {36.80} & {38.59} & 38.96 & {43.97} & {41.50} & 44.79 & 27.41 & 23.36 & 28.29 & 34.27 & 30.75 & 33.77 \\

    \rowcolor{purple!5} \textit{\small{Improvement}} & 
    - & \textcolor{blue}{\small{+7.95}} & \textcolor{blue}{\small{+7.03}} & \textcolor{blue}{\small{+8.75}} & \textcolor{blue}{\small{+9.15}} & \textcolor{blue}{\small{+8.63}} &\textcolor{blue}{\small{+10.94}} & \textcolor{blue}{\small{+6.17}} & \textcolor{blue}{\small{+5.45}} & \textcolor{blue}{\small{+8.71}} & \textcolor{blue}{\small{+7.85}} & \textcolor{blue}{\small{+6.77}} & \textcolor{blue}{\small{+8.57}}  \\
    \midrule
    DGCNN~\cite{wang2019dynamic} & {1.43~M} & 31.55 & 28.94 & 30.25 & 42.71 & 37.24 & 39.98 & 23.99 & 19.10 & 21.55 & 34.93 & 28.10 & 31.52 \\
    ProtoNet~\cite{garcia2017few} & 0.27~M & 33.92 & 30.95 & 32.44 & 45.34 & 42.01 & 43.68 & 28.47 & 26.13 & 27.30 & 37.36 & 34.98 & 36.17 \\ 
    
    MPTI~\cite{zhao2021few} & 0.29~M & 39.27 & 36.14 & 37.71 & 46.90 & 43.59 & 45.25 & 29.96 & 27.26 & 28.61 & 38.14 & 34.36 & 36.25 \\ 
    
    AttMPTI~\cite{zhao2021few} & 0.37~M & 42.55 & 40.83 & 41.69 & 54.00 & 50.32 & 52.16 & 35.23 & 30.72 & 32.98 & 46.74 & 40.80 & 43.77 \\ 
    BFG~\cite{mao2022bidirectional} & - & 42.15 & 40.52 & 41.34 & 51.23 & 49.39 & 50.31 & 34.12 & 31.98 & 33.05 & 46.25 & 41.38 & 43.82 \\
    
    2CBR~\cite{zhu2023cross} & 0.35~M & 50.73 & 47.66 & 49.20 & 52.35 & 47.14 & 49.75 & 47.00 & 46.36 & 46.68 & 45.06 & 39.47 & 42.27 \\ 
    PAP3D~\cite{he2023prototype} & 2.45~M & 57.08 & 55.94 & 56.51 & 64.55 & 59.64 & 62.10 & 55.27 & 55.60 & 55.44 & 59.02 & 53.16 & 56.09 \\
    \midrule
    
    \textbf{Seg-PN} & \textbf{0.24~M} & \textbf{63.15} & \textbf{64.32} & \textbf{63.74} & \textbf{67.08} & \textbf{69.05} & \textbf{68.07} & \textbf{61.80} & \textbf{65.34} & \textbf{63.57} & \textbf{62.94} & \textbf{68.26} & \textbf{65.60} \\
    \textit{\small{Improvement}} & 
    - & \textcolor{blue}{\small{+6.07}} & \textcolor{blue}{\small{+8.38}} & \textcolor{blue}{\small{+7.23}} & \textcolor{blue}{\small{+2.53}} & \textcolor{blue}{\small{+9.41}} &\textcolor{blue}{\small{+5.97}} & \textcolor{blue}{\small{+6.53}} & \textcolor{blue}{\small{+9.74}} & \textcolor{blue}{\small{+8.13}} & \textcolor{blue}{\small{+3.92}} & \textcolor{blue}{\small{+15.10}} & \textcolor{blue}{\small{+9.51}}  \\
    
\bottomrule
\end{tabular}
\end{adjustbox}
\caption{\textbf{Few-shot Results (\%) on ScanNet.} $S_i$ denotes the split $i$ is used for testing, and $Avg$ is their average mIoU. The shaded rows represent non-parametric methods. `Param.' represents the total number of learnable parameters of each method.}
\label{table:scannet_iou}
\vspace{-0.3cm}
\end{table*}

\begin{table}[t]
\centering
\begin{adjustbox}{width=0.997\linewidth}
	\begin{tabular}{lccccc}
	\toprule
		\makecell*[c]{\textbf{Method}} & \textbf{mIoU} & \textbf{Param.} & \makecell*[c]{\textbf{Pre-train}\\\textbf{Time}} & \makecell*[c]{\textbf{Episodic}\\\textbf{Train}} & \makecell*[c]{\textbf{Total}\\\textbf{Time}} \\
		\cmidrule(lr){1-1} \cmidrule(lr){2-2} \cmidrule(lr){3-3} \cmidrule(lr){4-6}
	DGCNN~\cite{wang2019dynamic} & 36.34 & 0.62~M & 4.0~h & 0.8~h & 4.8~h \\
    AttMPTI~\cite{zhao2021few} & 53.77 &  0.37~M & 4.0~h & 5.5~h & 9.5~h \\
    2CBR~\cite{zhu2023cross} & 55.89 & 0.35~M & 6.0~h & {0.2~h} & 6.2~h \\
    PAP3D~\cite{he2023prototype} & 59.45 & 2.45~M & 3.6~h & 1.1~h & 4.7~h \\
    \cmidrule(lr){1-6}
        
    \rowcolor{purple!5} \textbf{Seg-NN} & 49.45 & \textbf{0.00~M} & \textbf{0.0~h} & \textbf{0.0~h} & \textbf{0.0~h} \\
  
    \textbf{Seg-PN} & \textbf{64.84} & {0.24~M} & \textbf{0.0~h} & 0.5~h & {0.5~h}\\
    
	\bottomrule
	\end{tabular}
\end{adjustbox}
\caption{\textbf{Performance (\%) and Efficiency Comparison on S3DIS}. Train Time and Test Speed (episodes/second) are tested on one NVIDIA A6000 GPU. We calculate the total time of pre-training and episodic training. We report the accuracy under 2-way-1-shot settings on $S_0$ split.}
\label{table:efficiency_comparison}
\vspace{-0.3cm}
\end{table}

\subsection{Ablation Study}
\label{sec:ablation}
\vspace{-0.1cm}
We conduct extensive ablation experiments to reveal the roles of different designs. By default, we perform 2-way-1-shot experiments on the $S_0$ split of S3DIS dataset.

\begin{figure}[t]
\centering
\includegraphics[width=0.47\textwidth]{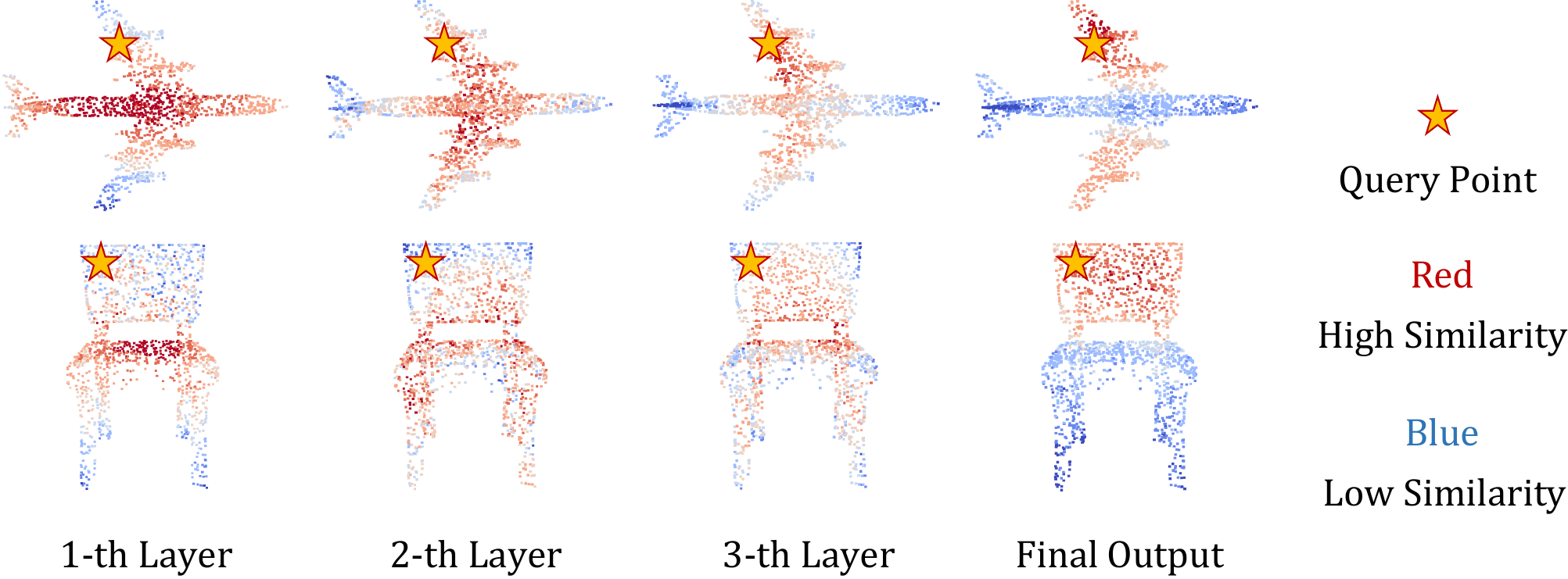}
\vspace{-0.1cm}
\caption{\textbf{Efficacy of the Hierarchical Structure.} We visualize the feature similarity between the ``query point'' and other points.}
\label{fig:hierarchical}
\vspace{-3pt}
\end{figure}

\vspace{-0.3cm}
\paragraph{Ablation for Seg-NN.} We first investigate the effect of different numbers of manipulation layers in Tab. \ref{table:ablation_number_layer}. We observe that Seg-NN and Seg-PN achieve their best performance with three and four layers respectively, indicating that Seg-PN prefers to learn deeper features. To demonstrate the efficacy of this hierarchical structure, we visualize the feature similarity map of each layer in Fig. \ref{fig:hierarchical}. The similarity map suggests our hierarchical design can gradually capture discriminative point-level features and the integration of multi-layer features can effectively model local shape characteristics.

\begin{table}[t]
\centering
\begin{adjustbox}{width=0.97\linewidth}
	\begin{tabular}{c c c cc}
	\toprule
		\makecell*[c]{\textbf{Backbone}} & \textbf{Projection} & \makecell*[c]{\textbf{QUEST}} & \makecell*[c]{\textbf{mIoU}} &\makecell*[c]{\textbf{Parameters}} 
		\\
		\cmidrule(lr){1-1} \cmidrule(lr){2-3} 
        \cmidrule(lr){4-5} 
   
	    \multirow{2}{*}{Point-NN~\cite{zhang2023parameter}}  & $\checkmark$ & \ding{55} & 46.31 & 1.20~M \\

          & \ding{55} & $\checkmark$ & 55.28 & 1.28~M \\
        Point-PN~\cite{zhang2023parameter}  & \ding{55} & $\checkmark$ & 54.74 & 3.97~M \\
        DGCNN~\cite{wang2019dynamic}  & \ding{55} & $\checkmark$ & 55.86 & 0.42~M \\

        \multirow{2}{*}{\textbf{Seg-NN}}  & $\checkmark$ & \ding{55} & 50.06 & 0.20~M \\
    
          & \ding{55}  & $\checkmark$ & \textbf{64.84} & 0.24~M \\
      
	\bottomrule
	\end{tabular}
\end{adjustbox}
\vspace{-0.1cm}
\caption{\textbf{Effect of Backbones and the QUEST Module on Seg-PN.} We combine different backbones with the QUEST module to demonstrate the efficacy of the proposed non-parametric encoder. We also substitute the QUEST module with a simple linear projection layer (`Projection') to verify its effect.}
\label{tab:quest_backbone}
\vspace{-0.3cm}
\end{table}

\vspace{-0.3cm}
\paragraph{Ablation on Seg-PN.} First, we investigate the role of the backbone and the QUEST module in Seg-PN in Tab. \ref{tab:quest_backbone}. By substituting our non-parametric encoder with other backbones, including Point-NN and pre-trained DGCNN and Point-PN (following \cite{he2023prototype}), we observe a performance drop, which highlights the effectiveness of our encoder in capturing 3D geometries. In addition, replacing the QUEST module with a projection layer also significantly impairs the performance. Tab. \ref{tab:quest_backbone} proves that combining our non-parametric encoder with the QUEST module can remarkably improve prediction. Then, we examine different designs of the QUEST module in Tab. \ref{table:ablation_QUEST}, including the roles of the self- and cross-correlation. We observe that using only self-correlation achieves lower performance than using cross-correlation, which indicates that the interaction between support and query sets is more important than only considering their self-correlation. Finally, utilizing both correlations allows for a higher performance. 

\begin{table}[t!]
\centering
\begin{adjustbox}{width=0.78\linewidth}
	\begin{tabular}{cccccc}
	\toprule
	\makecell*[c]{\textbf{Layers}} & {1} & {2} & {3} & {4} & {5}\\
    \cmidrule(lr){1-1} \cmidrule(lr){2-6}
    \specialrule{0em}{1pt}{1pt}
	Seg-NN & 36.43 & 42.38 & \textbf{49.45} & 46.19 & 43.18 \\
    \specialrule{0em}{1pt}{1pt}
    Seg-PN & 58.35 & 62.23 & 64.84 & \textbf{65.92} & 63.98 \\
    \specialrule{0em}{1pt}{1pt}
	\bottomrule
	\end{tabular}
\end{adjustbox}
\vspace{-0.1cm}
\caption{\textbf{Ablation for Number of Layers in Seg-NN Encoder.} }
\label{table:ablation_number_layer}
\vspace{-0.2cm}
\end{table}

\begin{table}[t]
\centering
\begin{adjustbox}{width=0.88\linewidth}
\begin{tabular}{l|ccc|ccc}
\toprule

    \multirow{2}{*}{\textbf{QUEST}} & \multicolumn{3}{c|}{\textbf{Two Way}} & \multicolumn{3}{c}{\textbf{Three Way}} \\ \cline{2-7}

    & $S_0$ & $S_1$ & $Avg$ & $S_0$ & $S_1$ & $Avg$ \\  \midrule
    
    w/o & {47.32} & {50.05} & 48.68 & 38.46 & 40.66 & 39.56 \\

    Self & 49.05 & 53.47 & 51.26 & {39.99} & 42.86 & 41.42 \\
    
    Cross & 62.72 & 67.16 & 64.94 & 59.73 & 61.49 & 60.61 \\

    QUEST & \textbf{64.84} & \textbf{67.98} & \textbf{66.41} & \textbf{60.12} & \textbf{63.22} & \textbf{61.67} \\
\bottomrule
\end{tabular}
\end{adjustbox}
\vspace{-0.1cm}
\caption{\textbf{Ablation for the QUEST Module} in Seg-PN. We report the results (\%) under 2/3-way-1-shot settings. `Self' and `Cross' represent self- and cross-correlation, respectively.}
\label{table:ablation_QUEST}
\vspace{-0.3cm}
\end{table}

\section{Additional Tasks}
To exhibit our generalization ability, we validate the Seg-NN encoder on other 3D tasks, including supervised classification, few-shot classification, and part segmentation.

\subsection{Training-free Classification}

We employ the Seg-NN encoder with minor modifications to encode each point cloud into a discriminative representation. Specifically, we discard the upsampling layers in the Seg-NN encoder and apply global maximum pooling to the embeddings obtained from the 3-th manipulation layer. This results in a global representation for each point cloud. To classify each point cloud, we refer to Point-NN \cite{zhang2023parameter} to adopt a similarity-matching scheme.

\vspace{-0.4cm}
\paragraph{Performance} We adopt ModelNet40~\cite{wu20153d} and ScanObjectNN~\cite{uy2019revisiting} datasets to evaluate the classification performance.
We compare Seg-NN with Point-NN~\cite{zhang2023parameter} in Tab. \ref{table:classification}. From the table, our method sets the new state-of-the-art for training-free classification on ModelNet40 and achieves comparable results on ScanObjectNN. In terms of inference speed, our method outperforms Point-NN by around +50\%.

\subsection{Training-free Few-shot Classification} 
We follow \cite{sharma2020self} to use ModelNet40 dataset \cite{wu20153d}. We conduct the $N$-way-$K$-shot classification tasks, $N \in \{5, 10 \}$ and $K \in \{10, 20 \}$. We randomly sample $N$ classes and $K$ point clouds for each class as the support set. For the query set, we pick 20 unseen objects from each of the $N$ classes. We randomly repeat 10 times and report the average results. 

\vspace{-0.4cm}
\paragraph{Performance} From Tab. \ref{table:fewshot_cls}, we observe that Seg-NN achieves the highest performance and significantly outperforms all existing parametric models, e.g., DGCNN~\cite{wang2019dynamic} and PointCNN~\cite{li2018pointcnn}.

\subsection{Training-free Part Segmentation.} 
We extend our non-parametric 3D encoder to the part segmentation task. In detail, we utilize the Seg-NN encoder to extract prototypes for each part category and employ the same similarity-based method to segment 3D objects into pre-defined part classes. 

\begin{table}[t]
\centering
\begin{adjustbox}{width=0.997\linewidth}
	\begin{tabular}{lccc|cc}
	\toprule
		\makecell*[c]{\textbf{Method}} & {MN40} & {SONN} & {Speed} & SNPart & {Speed} \\
		\cmidrule(lr){1-1}  \cmidrule(lr){2-4}
  \cmidrule(lr){5-6}
	    
        Point-NN~\cite{zhang2023parameter} & {81.8} & \textbf{64.9} & {209} & 70.4 & 52 \\
        
        \textbf{Seg-NN} & \textbf{84.2} & {64.4} & \textbf{306} & \textbf{73.2} & \textbf{88} \\
    
	\bottomrule
	\end{tabular}
\end{adjustbox}
\caption{\textbf{Training-free Performance (\%) and Efficiency}. Inference speed (samples/second) is tested on one NVIDIA A6000 GPU. We report the classification accuracy (\%) on ModelNet40 ('MN40') \cite{wu20153d} and ScanObjectNN ('SONN') \cite{uy2019revisiting} datasets, and mIoU (\%) of segmentation on ShapeNetPart (`SNPart')~\cite{yi2016scalable}.}
\label{table:classification}
\end{table}

\vspace{-0.4cm}
\paragraph{Performance} ShapeNetPart \cite{yi2016scalable} is a commonly used dataset for object-level part segmentation. From Tab. \ref{table:classification}, our method outperforms Point-NN by +2.8\% in segmentation performance and +60\% in inference speed.

\begin{table}[t]
\centering
\begin{adjustbox}{width=0.93\linewidth}
	\begin{tabular}{lc c c c }
	\toprule
        \makecell*[c]{\multirow{2}*{\textbf{Method}}} &\multicolumn{2}{c}{\textbf{5-way}} &\multicolumn{2}{c}{\textbf{10-way}} \\
        \cmidrule(lr){2-3} \cmidrule(lr){4-5}
        
        &10-shot &20-shot &10-shot &20-shot\\
        \cmidrule(lr){1-1} \cmidrule(lr){2-3}
        \cmidrule(lr){4-5}
		DGCNN~\cite{wang2019dynamic} &31.6 &40.8  &19.9 &16.9\\
        PointNet++~\cite{qi2017pointnet++}  &38.5 &42.4 &23.0 &18.8\\
	    
        3D-GAN~\cite{wu2016learning} &55.8 &65.8  &40.3 &48.4\\
	    PointCNN~\cite{li2018pointcnn} &65.4 &68.6  &46.6 &50.0\\

        Sharma et. al~\cite{sharma2020self} & 63.2 & 68.9 & 49.1 & 50.1 \\
     
        \cmidrule(lr){1-1} \cmidrule(lr){2-3}
        \cmidrule(lr){4-5}
        
        Point-NN~\cite{zhang2023parameter}\vspace{0.05cm} &88.8 &90.9  &79.9 &84.9\\

        \textbf{Seg-NN}\vspace{0.05cm} & \textbf{89.7} & \textbf{91.0}  & \textbf{81.7} & \textbf{86.1} \\
	
	\bottomrule
	\end{tabular}
\end{adjustbox}
\caption{\textbf{Few-shot Classification Results}  on ModelNet40~\cite{wu20153d}.}
\label{table:fewshot_cls}
\vspace{-5pt}
\end{table}

\section{Conclusion}

We propose a non-parametric framework, Seg-NN, and a parametric variant, Seg-PN, for few-shot point cloud semantic segmentation. Based on trigonometric PEs, Seg-NN introduces no learnable parameters and discards any training. Seg-NN achieves comparable performance with existing parametric methods and only causes minimal resource consumption. Furthermore, Seg-PN introduces the QUEST module to overcome prototype bias and achieves remarkable improvement compared to existing approaches. Importantly, Seg-PN does not require pre-training, thus simplifying the pipeline of 3D few-shot segmentation and saving significant time. For future work, we will explore the application of non-parametric encoders to more 3D tasks and improve the performance and generalization ability.
{
    \small
    \bibliographystyle{ieeenat_fullname}
    \bibliography{main}
}

\clearpage
\setcounter{page}{1}
\maketitlesupplementary

In this material, we first investigate the related literature about this work. Then, we provide detailed descriptions of the experimental setup and hyperparameters, along with additional framework details. At last, we present more ablation studies and further analysis.

\section{Related Work}
\label{sec:relatedwork}

\paragraph{Point Cloud Semantic Segmentation}aims to assign each point with the correct category label within a pre-defined label space. The early PointNet~\cite{qi2017pointnet}, PointNet++~\cite{qi2017pointnet++} establish the basic framework for learning-based 3D analysis. The follow-up works~\cite{wang2019dynamic,zhang2023parameter,zhao2021point,zhang2023learning,guo2023joint,zhang2022point,chen2023pimae} further improve the 3D representation and segmentation performance. Despite this, these methods are data-hungry and require extensive additional labeled data to be fine-tuned for unseen classes. To address this issue, a series of 3D few-shot learning methods have been proposed~\cite{mao2022bidirectional, zhang2023few, Xu2023Generalized, Liu2023PartSLIP, hu2023query, ning2023boosting, wang2023few}. AttMPTI~\cite{zhao2021few} extracts multiple prototypes from support-set features and predicts query labels in a transductive style. 2CBR~\cite{zhu2023cross} proposes cross-class rectification to alleviate the query-support domain gap. PAP-FZ3D~\cite{he2023prototype} jointly trains few-shot and zero-shot semantic segmentation tasks. All of these methods adopt the meta-learning strategy including both pre-training and episodic training stages. In this work, we propose more efficient solutions for 3D few-shot semantic segmentation. We first devise a non-parametric encoder to discard the time-consuming pre-training stage. Based on this, a parameter-free model, Seg-NN, and a parametric variant, Seg-PN, are proposed, which achieve competitive performance with minimal resources and simplify the traditional meta-learning pipelines.

\paragraph{Positional Encoding}(PE) projects a location vector into a high-dimensional embedding that can preserve spatial information and, at the same time, be learning-friendly for downstream algorithms \cite{mai2022review}. Transformer~\cite{vaswani2017attention} first utilizes PE to indicate the one-dimensional location of parallel input entries in a sequence, which is composed of trigonometric functions. Such trigonometric PE can encode both absolute and relative positions, and each of its dimensions corresponds to a predefined frequency and phase, which has also been employed for learning high-frequency functions~\cite{tancik2020fourier} and improving 3D rendering in NeRF~\cite{mildenhall2021nerf}. Some Transformers incorporate Gaussian random frequencies~\cite{kirillov2023segment}, and Mip-NeRF~\cite{barron2021mip} reduces aliasing artifacts in rendering by suppressing high frequencies. 
In 3D domains, RobustPPE \cite{zheng2023robust} adopts Gaussian random features for robust 3D classification. Point-NN~\cite{zhang2023parameter} is the first non-parametric model for shape classification which leverages basic trigonometric PEs to encode point coordinates for shape classification. 
In this work, we extend Point-NN to scene segmentation and utilize trigonometric PE to encode positional and color information, where manually designed filters are used for scene-level geometry encoding.

\section{Experimental Setup}

\paragraph{Dataset Split} 
S3DIS~\cite{armeni20163d} consists of 272 room point clouds from three different buildings with distinct architectural styles and appearances. We exclude the background clutter class and focus on 12 explicit semantic classes. ScanNet~\cite{dai2017scannet} comprises 1,513 point cloud scans from 707 indoor scenes, with 20 explicit semantic categories provided for segmentation. 
Tab. \ref{tab:data_split} lists the class names in the $S_0$ and $S_1$ splits of S3DIS and ScanNet datasets.

\paragraph{Hyperparameters} The Seg-NN encoder is frozen in all experiments. In the Seg-NN encoder, we sample 16 neighbor points with $k$-NN to build the neighborhood of the center point for both manipulation and upsampling layers. For Seg-PN, we first use a fully connected layer to refine the features extracted by the non-parametric encoder and then feed the refined features to the QUEST module. The fully connected layer consists of 2 linear projection operations and each linear projection is followed by a batch normalization~\cite{ioffe2015batch} and a rectified linear activation~\cite{glorot2011deep} function. The detailed structure of the fully connected layer is: `(BN+ReLU) + (Linear+BN+ReLU) + (Linear+BN+ReLU)', where `BN', `ReLU', and `Linear' represent batch normalization, rectified linear activation, and linear projection, respectively. We set the kernel size and stride of the local maximum pooling to 32 in the QUEST module. Since each point cloud contains $M=2048$ points, the local maximum pooling operation outputs $M'=64$ statistics for each point cloud. 

\vspace{-0.1cm}
\paragraph{Training Details} The proposed Seg-NN and Seg-PN are implemented using PyTorch. Seg-PN is trained on a GForce A6000 GPU. The meta-training is performed directly on $C_{train}$ split, using AdamW optimizer ($\beta_{1} = 0.9, \beta_{2} = 0.999$) to update the QUEST module of Seg-PN. The initial learning rate is set to 0.001 and halved every 7,000 iterations. In episodic training, each batch contains 1 episode, which includes a support set and a query set. The support set randomly selects $N$-way-$K$-shot point clouds and the query set randomly selects $N$ unseen samples.

\begin{table}[t]
\centering
\begin{adjustbox}{width=0.97\linewidth}
	\begin{tabular}{c|c|c}
	\toprule
	 & $S_0$ & $S_1$ \\
        \hline
		 \textbf{S3DIS} & \thead{beam, board, bookcase, \\ ceiling, chair, column} & \thead{door, floor, sofa, \\ table, wall, window} \\
		 \hline
         \textbf{ScanNet} & \thead{bathtub, bed, bookshelf, \\
        cabinet, chair, counter, \\ curtain, desk, door, \\floor} & \thead{other furniture, picture, \\refrigerator, show curtain,\\
        sink, sofa, table, \\toilet, wall, window} \\
		\specialrule{0em}{1pt}{1pt}
	\bottomrule
	\end{tabular}
\end{adjustbox}
\caption{\textbf{Seen and Unseen Classes Split} for S3DIS and ScanNet. We follow \cite{zhao2021few} to evenly assign categories to $S_0$ and $S_1$ splits.}
\label{tab:data_split}
\vspace{-0.2cm}
\end{table}

\section{Additional Ablation Study}

We conduct additional ablation experiments to reveal the roles of different detailed designs. By default, we still conduct experiments under 2-way-1-shot settings on the $S_0$ split of S3DIS dataset and use mIoU (\%) as criteria to evaluate the results of both Seg-NN and Seg-PN.

\subsection{Ablation for Seg-NN}
In this section, we mainly investigate different hyperparameters and designs of PEs, embedding manipulation layers, and upsampling layers.

\paragraph{Role of Position and Color Information} In Tab. \ref{table:ablation_position_color}, we exhibit more results to investigate the role of the position and color information. For Seg-NN, we observe that both position and color information are helpful for the segmentation, indicating that our model is capable of encoding geometries and integrating two types of information. However, in Seg-PN, employing colors hinders the prediction, which suggests that color information is not crucial for few-shot tasks and may lead to overfitting. This aligns with the observation of \cite{qian2022pointnext}, which randomly abandons color information during training to reduce overfitting.

\begin{table}[t!]
\centering
\begin{adjustbox}{width=0.99\linewidth}
\begin{tabular}{l|ccc|ccc}
\toprule
    \multirow{2}{*}{\textbf{mIoU}} & \multicolumn{3}{c|}{\textbf{Coordinate}} & \multicolumn{3}{c}{\textbf{+ Color}} \\ \cline{2-7}

    & $S_0$ & $S_1$ & $Avg$ & $S_0$ & $S_1$ & $Avg$ \\  \midrule

    Seg-NN & 48.81 & 49.04 & 48.93 & \textbf{49.45} & \textbf{49.60} & \textbf{49.53} \\

    Seg-PN & \textbf{67.47} & 66.78 & \textbf{67.13} & 64.84 & \textbf{67.98} & 66.41 \\
\bottomrule
\end{tabular}
\end{adjustbox}
\caption{\textbf{Ablation for Position and Color Information} under 2-way-1-shot settings on both $S_0$ and $S_1$ splits of S3DIS. We report Seg-NN and Seg-PN's results (\%).}
\label{table:ablation_position_color}
\end{table}

\begin{table}[t!]
\centering
\vspace{-2pt}
\begin{adjustbox}{width=0.99\linewidth}
	\begin{tabular}{cccccccc}
	\toprule
	\makecell*[c]{\textbf{$\theta$}} & {10} & {20} & 30 & {40} & {60} & {80} & {100}\\
    \cmidrule(lr){1-1} \cmidrule(lr){2-8}
    \specialrule{0em}{1pt}{1pt}
	Seg-NN & 49.12 & 49.38 & \textbf{49.45} & 47.87 & 44.76 & 43.54 & 40.16 \\
	\specialrule{0em}{1pt}{1pt}
    Seg-PN & 61.85 & 62.34 & \textbf{64.84} & 63.08 & 62.56 & 63.68 & 64.05 \\
	\specialrule{0em}{1pt}{1pt}
	\bottomrule
	\end{tabular}
\end{adjustbox}
\vspace{-2pt}
\caption{\textbf{Ablation for Parameter $\theta$ in PEs}.}
\label{table:ablation_theta}
\end{table}

\begin{table}[t]
\centering
\vspace{-4pt}
\begin{adjustbox}{width=0.99\linewidth}
	\begin{tabular}{cccccccc}
	\toprule
	\makecell*[c]{$d$} & {5} & {8} & {10} & {15} & {20} & {24} & 30 \\
        \cmidrule(lr){1-1} \cmidrule(lr){2-8}
        \specialrule{0em}{1pt}{1pt}
		\specialrule{0em}{1pt}{1pt}
         Seg-NN & 44.24 & 46.37 & 47.65 & 48.90 & \textbf{49.45} & 48.68 & 48.53 \\
         Seg-PN & 63.37 & 63.49 & \textbf{64.84} & 63.76 & 63.35 & 63.45 & 63.81 \\ 
		 \specialrule{0em}{1pt}{1pt}
	\bottomrule
	\end{tabular}
\end{adjustbox}
\caption{\textbf{Ablation on the Dimensionality of PEs}.}
\label{table:ablation_PEdim_d}
\vspace{-0.2cm}
\end{table}

\paragraph{Hyperparameters of PEs} \textbf{1) Parameter $\mathbf{\theta}$ in PE.} In the initial PE, we utilize $d$ log-linear spaced frequencies $\mathbf{u}=[u_1, ..., u_d]$ to project positions and colors into high-dimensional encodings, where $u_{i}=\theta^{i/d}$ with a base number $\theta$. In Tab. \ref{table:ablation_theta}, we explore the impact of $\theta$ on Seg-NN and Seg-PN, where $\theta \in \{10, 20, 30, 40, 60, 80, 100\}$. From the table, we observe that Seg-NN is more sensitive to $\theta$, while Seg-PN exhibits higher tolerance. In addition, Seg-NN prefers low $\theta$ values, and a larger $\theta$ will cause significant performance degradation. \textbf{2) Dimensionality of PEs.} We then examine the effect of the dimensionality of PEs. We sample $d$ frequencies to construct the PE. Tab. \ref{table:ablation_PEdim_d} presents the results with different frequency numbers $d$. We explore $d \in \{5, 8, 10, 15, 20, 24, 30\}$ and the corresponding dimensionality of PEs are $6d \in \{30, 48, 60, 90, 120, 144, 180\}$. We observe that $d=20$ and $10$ are the best choices for Seg-NN and Seg-PN, respectively. This suggests that reducing the dimension of PEs has the potential to impair the performance of Seg-NN, while Seg-PN can effectively learn shape representations from relatively lower-dimensional embeddings.

\paragraph{Embedding Manipulation}  \textbf{1) Different Distributions of Sampled Frequencies} In embedding manipulation layers, we sample frequencies $\mathbf{v}$ to generate the projection weights. In Tab. \ref{table:ablation_distribution}, we compare the effects of different frequency distributions. Totally three types of distribution are compared, Gaussian, Laplace, and uniform distributions. By comparison, we observe that the best performance is achieved when the sampled frequencies follow a Gaussian distribution. The reason behind this may be that both Laplace and uniform distribution contain more mid- and high-frequency information, thereby introducing excessive noises and redundancies into shape representation.    \textbf{2) Variance of Gaussian Distribution.} In Tab. \ref{table:ablation_variance}, we investigate the impact of the variance of the Gaussian distribution used for frequency sampling. A larger variance indicates more middle or high frequencies are exploited in feature extraction. We find that Seg-PN can learn useful information from higher frequencies to enhance performance, while  Seg-NN benefits more from low frequencies.

\paragraph{Scaling Factor $\gamma$ in the Segmentation Head} In the non-parametric segmentation head, we use $\varphi(x)=\exp(-\gamma(1-x))$ as an activation function, where $\gamma$ is a scaling factor. In this part, we explore the effect of different values of $\gamma$. Tab. \ref{table:ablation_gamma} presents the results of Seg-NN with different $\gamma$s. We experiment with $\gamma \in \{100, 300, 500, 700, 1000, 1200, 1500\}$ and observe that $\gamma \le 500$ guarantees more accurate prediction and $\gamma>1000$ causes a rapid performance drop.

\begin{table}[t]
\centering
\begin{adjustbox}{width=0.95\linewidth}
\begin{tabular}{ccc|cc}
\toprule
    \multicolumn{3}{c|}{\textbf{Frequency Distribution}} & \multirow{2}{*}{{Seg-NN}} & \multirow{2}{*}{{Seg-PN}}\\ \cline{1-3}

    Gaussian & Laplace & Uniform \\  \hline
    
    $\checkmark$ & & & \textbf{49.45} & \textbf{64.84} \\
    & $\checkmark$ &  & 45.21 & 63.21 \\
    & & $\checkmark$ & 49.35 & 62.49 \\
\bottomrule
\end{tabular}
\end{adjustbox}
\caption{\textbf{Ablation Study for Frequency Distribution} in embedding manipulation layers.}
\label{table:ablation_distribution}
\end{table}

\begin{table}[t!]
\centering
\begin{adjustbox}{width=0.92\linewidth}
	\begin{tabular}{ccccccc}
	\toprule
	\makecell*[c]{\textbf{Variance}} & {0.5} & {1} & {2} & {5} & {10} & {20}\\
    \cmidrule(lr){1-1} \cmidrule(lr){2-7}
    \specialrule{0em}{1pt}{1pt}
	Seg-NN & 48.37 & \textbf{49.45} & 49.43 & 49.02 & 47.53  & 46.13  \\
	\specialrule{0em}{1pt}{1pt}
    Seg-PN & 63.87 & 64.84 & 64.86 & \textbf{65.72} & 64.29 & 63.10  \\
	\specialrule{0em}{1pt}{1pt}
	\bottomrule
	\end{tabular}
\end{adjustbox}
\vspace{-0.1cm}
\caption{\textbf{Ablation for the Variance of the Gaussian Distribution} in frequencies sampling.}
\label{table:ablation_variance}
\vspace{-0.2cm}
\end{table}

\begin{table}[t!]
\centering
\begin{adjustbox}{width=0.97\linewidth}
	\begin{tabular}{cccccccc}
	\toprule
	\makecell*[c]{\textbf{$\gamma$}} & {100} & {300} & {500} & {700} & {1000} & {1200} & 1500\\
    \cmidrule(lr){1-1} \cmidrule(lr){2-8}
    \specialrule{0em}{1pt}{1pt}
	Seg-NN & 50.25 & 50.13 & \textbf{50.66} & 50.17 & 49.45 & 44.21 & 31.32
 \\
	\specialrule{0em}{1pt}{1pt}
	\bottomrule
	\end{tabular}
\end{adjustbox}
\vspace{-1pt}
\caption{\textbf{Ablation for Scaling Factor $\gamma$ in the Segmentation Head} of Seg-NN.}
\label{table:ablation_gamma}
\end{table}

\begin{table}[t]
\centering
\vspace{-4pt}
\begin{adjustbox}{width=0.97\linewidth}
	\begin{tabular}{ccccccc}
	\toprule
	\makecell*[c]{\textbf{Kernel Size}} & {8} & {16} & {24} & {32} & {40} & {48} \\
        \cmidrule(lr){1-1} \cmidrule(lr){2-7}
        \specialrule{0em}{1pt}{1pt}
		\specialrule{0em}{1pt}{1pt}
         Seg-PN & 59.67 & 64.87 & \textbf{66.06} & 64.84 & 64.75 & 63.90  \\ 
		 \specialrule{0em}{1pt}{1pt}
	\bottomrule
	\end{tabular}
\end{adjustbox}
\caption{\textbf{Ablation for the Kernel Size and Stride} of the local maximum pooling operation in the QUEST module.}
\label{table:ablation_kernelsize}
\vspace{-0.1cm}
\end{table}

\begin{table}[t]
\centering
\begin{adjustbox}{width=0.85\linewidth}
\begin{tabular}{l|cc|cc}
    \toprule
    {\textbf{Source}} & \multicolumn{2}{c|}{\textbf{S3DIS}} & \multicolumn{2}{c}{\textbf{ScanNet}} \\ \cmidrule(lr){1-1} \cmidrule(lr){2-3} \cmidrule(lr){4-5}
    {\textbf{Target}} & S3DIS & ScanNet & S3DIS & ScanNet \\  \midrule
    Seg-NN & 59.4 & 43.9 & 59.4 & 43.9 \\
    Seg-PN & \textbf{67.6} & \textbf{64.6} & \textbf{63.3} & \textbf{67.0} \\ \midrule
    DGCNN & 56.6 & 44.8 & 49.4 & 42.7 \\
    AttMPTI & 61.6 & 46.3 & 49.7 & 54.0  \\
    2CBR & 63.5 & 49.6 & 54.9 & 52.3  \\
    PAP3D & 65.4 & 52.3 & 57.0 & 64.5 \\
\bottomrule
\end{tabular}
\end{adjustbox}
\vspace{-7pt}
\caption{\textbf{Transferability among datasets.} We report the results under 2-way-5-shot settings on the $S_0$ split.}
\label{table:domain_gap}
\vspace{-0.1cm}
\end{table}

\subsection{Ablation for Seg-PN}
\paragraph{Pooling Operation in the QUEST Module} In the QUEST module, we use local maximum pooling to obtain $M'$ statistics for each point cloud. We mainly explore two hyperparameters: the kernel size and stride of the local maximum pooling, the values of which are set to be the same. Tab. \ref{table:ablation_kernelsize} presents the effect of different kernel sizes. We observe that the best performance is achieved when the kernel size is 24, though the experiments in the main paper are conducted with a kernel size of 32.

\subsection{More Analysis}

\paragraph{Reduction of `seen'/`unseen' domain gap.}
\textbf{1)} We have shown that Seg-NN can reduce the `seen'/`unseen' domain gap in the main paper's Fig. 2 (a). We further extend the experiments in Fig. \ref{fig:miou}, where the mIoU difference between `seen' and `unseen' classes by our Seg-NN and Seg-PN is much smaller (DGCNN's 38\% vs our 3.1\% and 12.0\% on average).
\textbf{2)} We also show the t-SNE visualization in Fig. \ref{fig:tsne}, where the 3D features by Seg-NN are more discriminative among `unseen' classes than DGCNN. This indicates the `seen'/`unseen' semantic gap can be significantly alleviated by our encoder. The DGCNN in this experiment is trained on seen classes and tested on both seen and unseen classes. 

\paragraph{Transferability among different datasets.} In addition to the domain gap between the support set and query set, a natural extension is to investigate the transferability of our non-parametric model across different data domains. In Tab. \ref{table:domain_gap}, we present the transferring performance between S3DIS~\cite{armeni20163d} and ScanNet~\cite{dai2017scannet} datasets. We train models on the source dataset and then utilize the target dataset to evaluate the model. As shown in the table, \textit{even trained on S3DIS, our Seg-PN can attain the best ScanNet performance} compared to all existing methods. The results demonstrate our superior cross-dataset generalization capacity.

\section{Visualization}
\begin{figure}[t]
\centering
\includegraphics[width=0.455\textwidth]{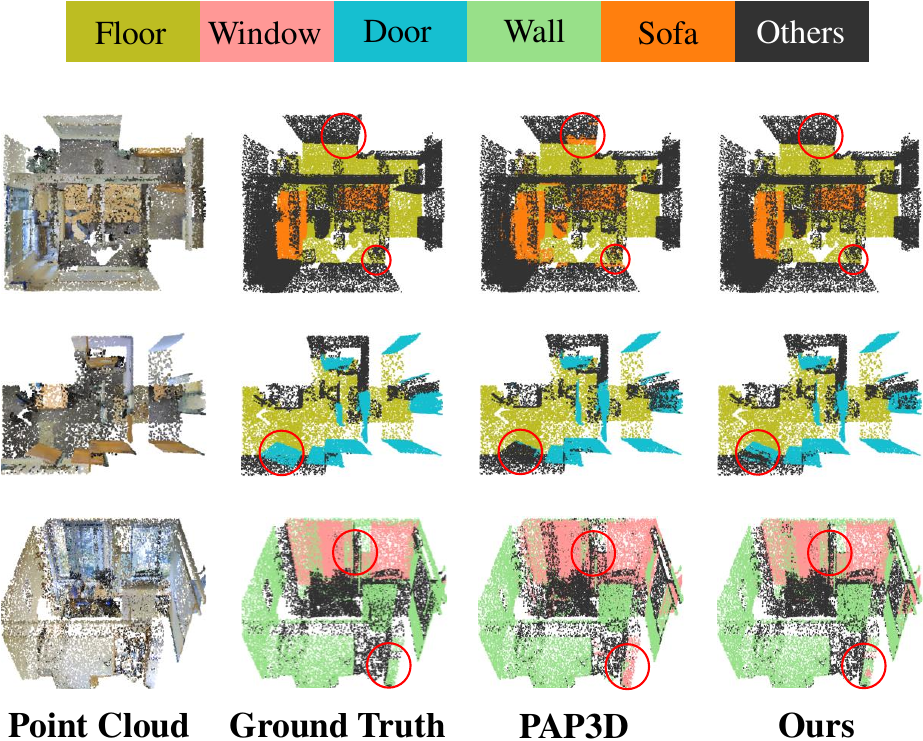}
\caption{\textbf{Visualization of Results} on S3DIS dataset. We compare Seg-PN's results with the SOTA PAP3D model.}
\label{fig:eg_s3dis}
\end{figure}

\begin{figure}[t]
\vspace{6pt}
\centering
\includegraphics[width=7.0cm]{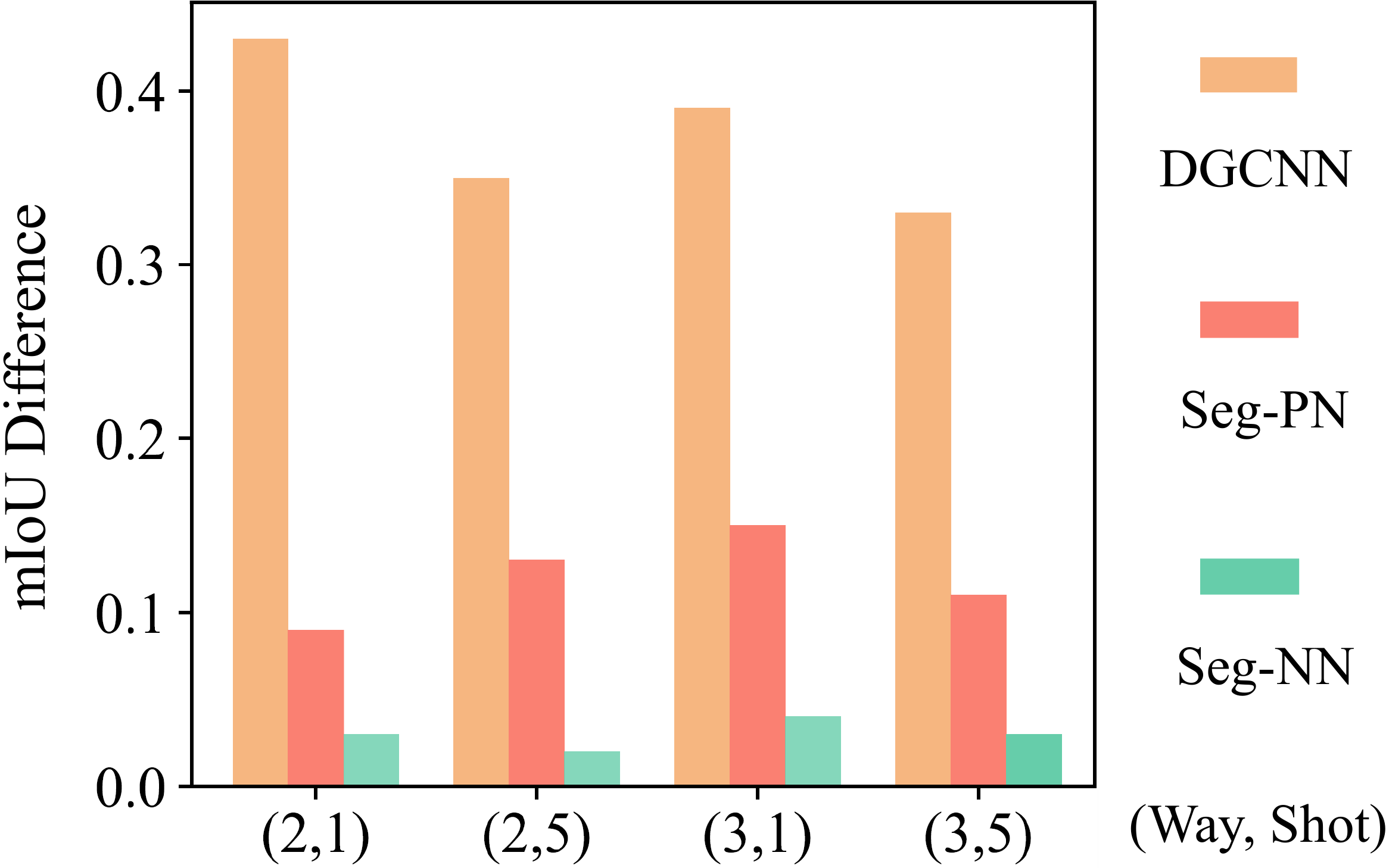}
\caption{\textbf{`Seen'/`unseen' Performance Gap.} We compare the performance difference of segmentation on the S3DIS dataset, where DGCNN shows a large performance difference between seen and unseen classes.}
\label{fig:miou}
\vspace{7cm}
\end{figure}

\begin{figure}[t]
\centering
\includegraphics[width=0.463\textwidth]{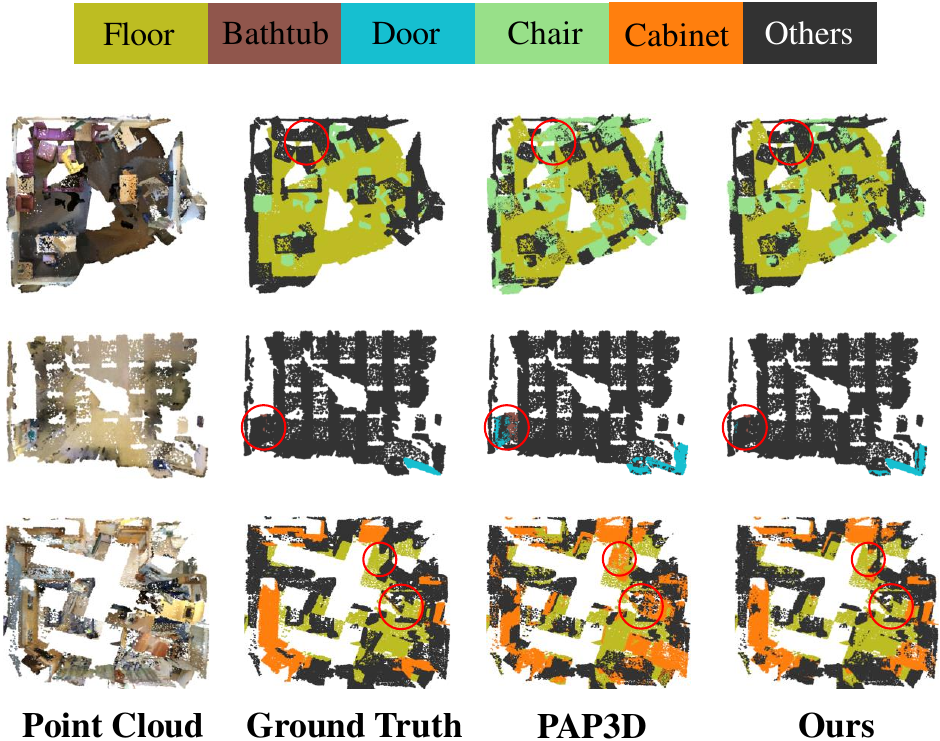}
\caption{\textbf{Visualization of Results} on ScanNet dataset. We compare Seg-PN's results with the SOTA PAP3D model.}
\label{fig:eg_scannet}
\end{figure}

\begin{figure}[t]
\vspace{-2pt}
\centering
\includegraphics[width=7.85cm]{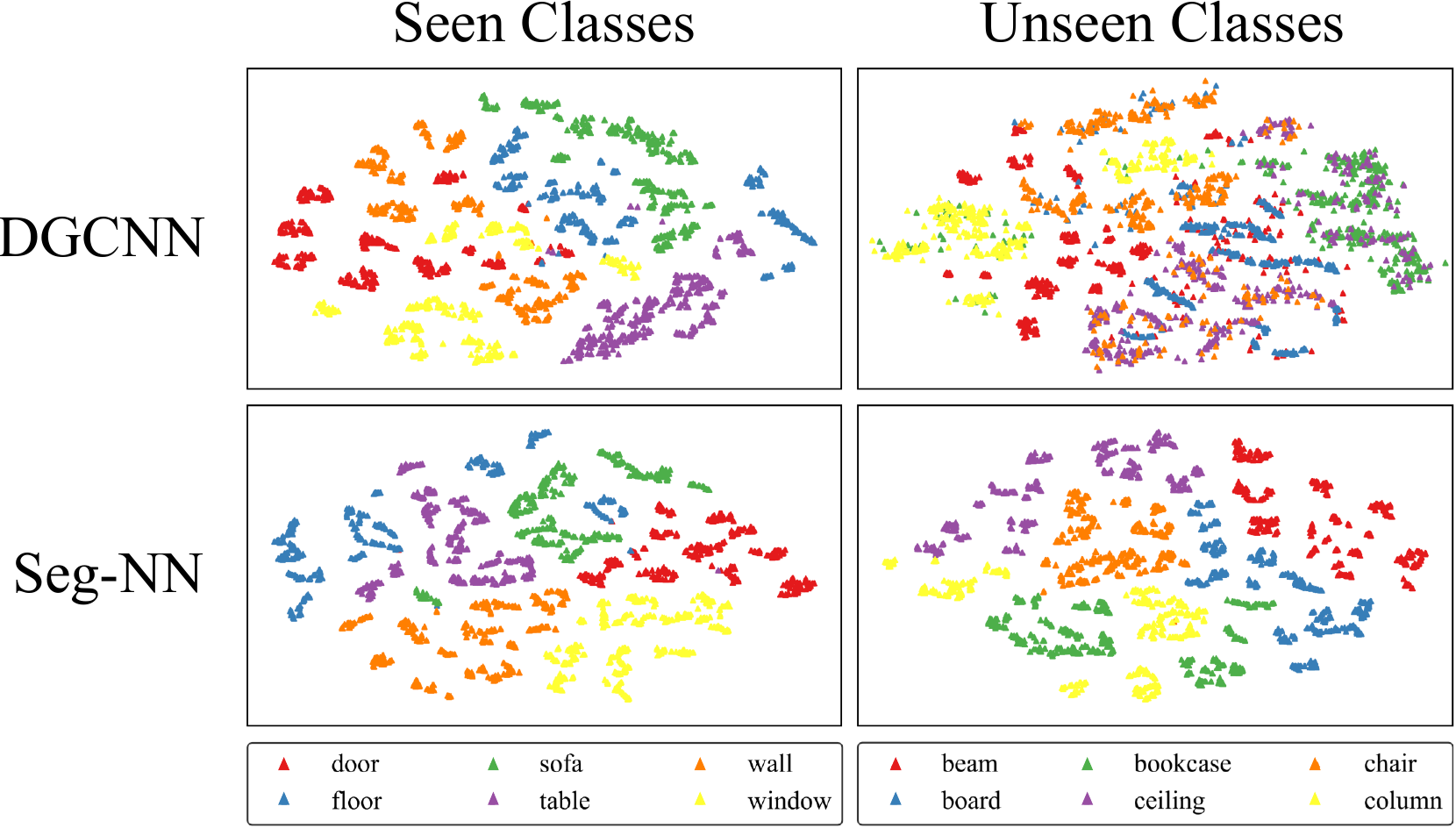}
\vspace{3pt}
\caption{\textbf{t-SNE Visualization of Features} on S3DIS, which suggests the non-parametric Seg-NN can extract discriminative embeddings for both seen and unseen classes.}
\label{fig:tsne}
\vspace{7.4cm}
\end{figure}

We present several qualitative results of 2-way-1-shot tasks in Fig. \ref{fig:eg_s3dis} and Fig. \ref{fig:eg_scannet}. Seg-PN achieves better segmentation than the existing SOTA, PAP3D~\cite{he2023prototype}, which demonstrates the effectiveness of Seg-PN. 
It is worth noting that due to sparse sampling in certain regions of the rooms, some ScanNet rooms may appear incomplete, as shown in Fig. \ref{fig:eg_scannet}. All rooms are presented in a top-down view.

\end{document}